%% file: main.tex
\documentclass{article}

\input{preamble}

\usepackage[final]{neurips_2025}

\usepackage[utf8]{inputenc} 
\usepackage[T1]{fontenc}    
\usepackage{hyperref}       
\usepackage{url}            
\usepackage{booktabs}       
\usepackage{amsfonts}       
\usepackage{nicefrac}       
\usepackage{microtype}      
\usepackage{xcolor}         
\usepackage{graphicx}
\usepackage{caption}
\usepackage{enumitem}
\usepackage{pifont}
\usepackage{mathtools}
\usepackage{dblfloatfix}
\usepackage{colortbl}  
\usepackage{xcolor}    
\usepackage{xspace}
\usepackage{wrapfig}
\usepackage{multirow}
\usepackage{hyperref}
\usepackage{algorithmic}
\usepackage[ruled,vlined]{algorithm2e}
\SetKwInOut{Input}{Input}
\SetKwInOut{Output}{Output}
\SetKwInOut{Require}{Require}
\usepackage{lineno}
\usepackage{etoc} 

\usepackage{authblk}

\newcommand{\venueTT}[1]{{$_{\texttt{\text{#1}}}$}}

\newcommand{\cmark}{\text{\ding{51}}}
\newcommand{\xmark}{\textcolor[rgb]{0.7,0.7,0.7}{\text{\ding{55}}}}

\definecolor{mypink}{HTML}{FB2E99}
\definecolor{myred}{RGB}{255,150,150}       
\definecolor{myorange}{RGB}{255,200,120}    
\definecolor{myyellow}{RGB}{255,255,160}    

\title{EGGS: Exchangeable 2D/3D Gaussian Splatting for Geometry-Appearance Balanced Novel View Synthesis}

\author[ ]{Yancheng Zhang}
\author[ ]{Guangyu Sun}
\author[ ]{Chen Chen}

\affil[ ]{Institute of Artificial Intelligence, University of Central Florida}
\affil[ ]{\texttt{\{yczhang, guangyu.sun, chen.chen\}@ucf.edu}}

\begin{document}

\maketitle

\begin{figure*}[h]
    \vspace{-0.3in}
    \centering
    \includegraphics[width=1\linewidth]{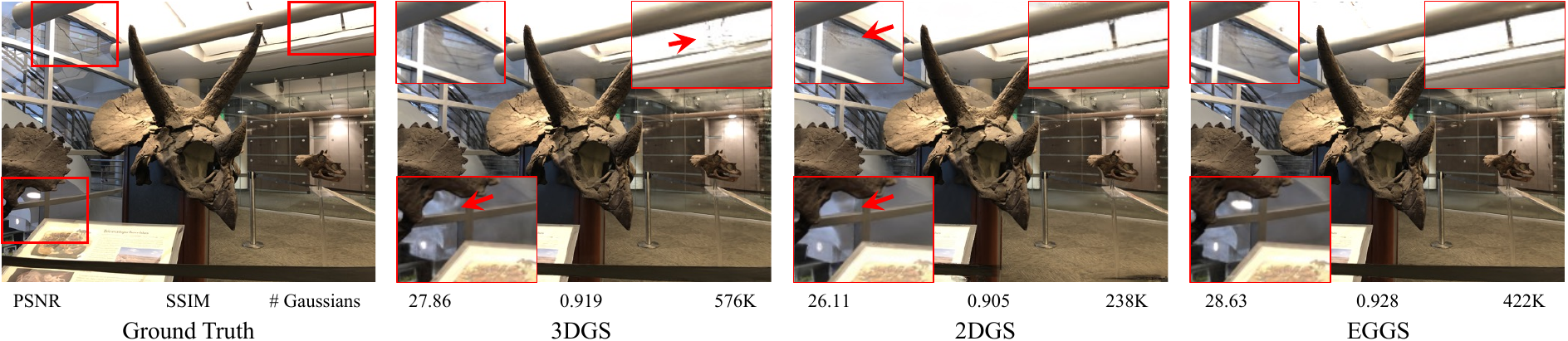}
    \caption{Comparison of 3DGS, 2DGS, and our \method. While 3DGS achieves high-fidelity appearance, it often produces inaccurate geometry, with imprecise surfaces and blurred edges. 2DGS improves geometric consistency across views but suffers from reduced appearance quality due to over-smoothed surfaces and loss of detail. In contrast, \method employs an exchangeable hybrid Gaussian representation that achieves both accurate geometry and high-quality appearance.}
    \label{fig:teaser}
    \vspace{-0.1in}
\end{figure*}

\input{sec/0_abstract}    
\input{sec/1_intro}
\input{sec/2_background}
\input{sec/3_method}

\input{sec/4_expt}

\input{sec/5_conclusion}


\bibliography{3d}
\bibliographystyle{unsrt}


\clearpage
\appendix
\section*{Appendix}

\input{sec/6_appendix}

\clearpage
\newpage
\section*{NeurIPS Paper Checklist}
\begin{enumerate}

\item {\bf Claims}
    \item[] Question: Do the main claims made in the abstract and introduction accurately reflect the paper's contributions and scope?
    \item[] Answer: \answerYes{} 
    \item[] Justification: The main claims in the abstract and introduction clearly align with the paper's contributions and scope.
    \item[] Guidelines:
    \begin{itemize}
        \item The answer NA means that the abstract and introduction do not include the claims made in the paper.
        \item The abstract and/or introduction should clearly state the claims made, including the contributions made in the paper and important assumptions and limitations. A No or NA answer to this question will not be perceived well by the reviewers. 
        \item The claims made should match theoretical and experimental results, and reflect how much the results can be expected to generalize to other settings. 
        \item It is fine to include aspirational goals as motivation as long as it is clear that these goals are not attained by the paper. 
    \end{itemize}

\item {\bf Limitations}
    \item[] Question: Does the paper discuss the limitations of the work performed by the authors?
    \item[] Answer: \answerYes{} 
    \item[] Justification: The limitations are explicitly discussed in the discussion section, providing a clear understanding of the work's boundaries.
    \item[] Guidelines:
    \begin{itemize}
        \item The answer NA means that the paper has no limitation while the answer No means that the paper has limitations, but those are not discussed in the paper. 
        \item The authors are encouraged to create a separate "Limitations" section in their paper.
        \item The paper should point out any strong assumptions and how robust the results are to violations of these assumptions (e.g., independence assumptions, noiseless settings, model well-specification, asymptotic approximations only holding locally). The authors should reflect on how these assumptions might be violated in practice and what the implications would be.
        \item The authors should reflect on the scope of the claims made, e.g., if the approach was only tested on a few datasets or with a few runs. In general, empirical results often depend on implicit assumptions, which should be articulated.
        \item The authors should reflect on the factors that influence the performance of the approach. For example, a facial recognition algorithm may perform poorly when image resolution is low or images are taken in low lighting. Or a speech-to-text system might not be used reliably to provide closed captions for online lectures because it fails to handle technical jargon.
        \item The authors should discuss the computational efficiency of the proposed algorithms and how they scale with dataset size.
        \item If applicable, the authors should discuss possible limitations of their approach to address problems of privacy and fairness.
        \item While the authors might fear that complete honesty about limitations might be used by reviewers as grounds for rejection, a worse outcome might be that reviewers discover limitations that aren't acknowledged in the paper. The authors should use their best judgment and recognize that individual actions in favor of transparency play an important role in developing norms that preserve the integrity of the community. Reviewers will be specifically instructed to not penalize honesty concerning limitations.
    \end{itemize}

\item {\bf Theory assumptions and proofs}
    \item[] Question: For each theoretical result, does the paper provide the full set of assumptions and a complete (and correct) proof?
    \item[] Answer: \answerYes{} 
    \item[] Justification: The assumptions for each theoretical result are clearly stated, and the correctness is analyzed in detail in the appendix.
    \item[] Guidelines:
    \begin{itemize}
        \item The answer NA means that the paper does not include theoretical results. 
        \item All the theorems, formulas, and proofs in the paper should be numbered and cross-referenced.
        \item All assumptions should be clearly stated or referenced in the statement of any theorems.
        \item The proofs can either appear in the main paper or the supplemental material, but if they appear in the supplemental material, the authors are encouraged to provide a short proof sketch to provide intuition. 
        \item Inversely, any informal proof provided in the core of the paper should be complemented by formal proofs provided in appendix or supplemental material.
        \item Theorems and Lemmas that the proof relies upon should be properly referenced. 
    \end{itemize}

    \item {\bf Experimental result reproducibility}
    \item[] Question: Does the paper fully disclose all the information needed to reproduce the main experimental results of the paper to the extent that it affects the main claims and/or conclusions of the paper (regardless of whether the code and data are provided or not)?
    \item[] Answer: \answerYes{} 
    \item[] Justification: The method and algorithms are clearly explained, with parameters and experimental settings provided to ensure reproducibility of the main results.
    \item[] Guidelines:
    \begin{itemize}
        \item The answer NA means that the paper does not include experiments.
        \item If the paper includes experiments, a No answer to this question will not be perceived well by the reviewers: Making the paper reproducible is important, regardless of whether the code and data are provided or not.
        \item If the contribution is a dataset and/or model, the authors should describe the steps taken to make their results reproducible or verifiable. 
        \item Depending on the contribution, reproducibility can be accomplished in various ways. For example, if the contribution is a novel architecture, describing the architecture fully might suffice, or if the contribution is a specific model and empirical evaluation, it may be necessary to either make it possible for others to replicate the model with the same dataset, or provide access to the model. In general. releasing code and data is often one good way to accomplish this, but reproducibility can also be provided via detailed instructions for how to replicate the results, access to a hosted model (e.g., in the case of a large language model), releasing of a model checkpoint, or other means that are appropriate to the research performed.
        \item While NeurIPS does not require releasing code, the conference does require all submissions to provide some reasonable avenue for reproducibility, which may depend on the nature of the contribution. For example
        \begin{enumerate}
            \item If the contribution is primarily a new algorithm, the paper should make it clear how to reproduce that algorithm.
            \item If the contribution is primarily a new model architecture, the paper should describe the architecture clearly and fully.
            \item If the contribution is a new model (e.g., a large language model), then there should either be a way to access this model for reproducing the results or a way to reproduce the model (e.g., with an open-source dataset or instructions for how to construct the dataset).
            \item We recognize that reproducibility may be tricky in some cases, in which case authors are welcome to describe the particular way they provide for reproducibility. In the case of closed-source models, it may be that access to the model is limited in some way (e.g., to registered users), but it should be possible for other researchers to have some path to reproducing or verifying the results.
        \end{enumerate}
    \end{itemize}

\item {\bf Open access to data and code}
    \item[] Question: Does the paper provide open access to the data and code, with sufficient instructions to faithfully reproduce the main experimental results, as described in supplemental material?
    \item[] Answer: \answerYes{} 
    \item[] Justification: The evaluation primarily uses public datasets cited and explained in the experimental setting section, and the code is available in the anonymous repository with sufficient instructions for reproduction.
    \item[] Guidelines:
    \begin{itemize}
        \item The answer NA means that paper does not include experiments requiring code.
        \item Please see the NeurIPS code and data submission guidelines (\url{https://nips.cc/public/guides/CodeSubmissionPolicy}) for more details.
        \item While we encourage the release of code and data, we understand that this might not be possible, so “No” is an acceptable answer. Papers cannot be rejected simply for not including code, unless this is central to the contribution (e.g., for a new open-source benchmark).
        \item The instructions should contain the exact command and environment needed to run to reproduce the results. See the NeurIPS code and data submission guidelines (\url{https://nips.cc/public/guides/CodeSubmissionPolicy}) for more details.
        \item The authors should provide instructions on data access and preparation, including how to access the raw data, preprocessed data, intermediate data, and generated data, etc.
        \item The authors should provide scripts to reproduce all experimental results for the new proposed method and baselines. If only a subset of experiments are reproducible, they should state which ones are omitted from the script and why.
        \item At submission time, to preserve anonymity, the authors should release anonymized versions (if applicable).
        \item Providing as much information as possible in supplemental material (appended to the paper) is recommended, but including URLs to data and code is permitted.
    \end{itemize}

\item {\bf Experimental setting/details}
    \item[] Question: Does the paper specify all the training and test details (e.g., data splits, hyperparameters, how they were chosen, type of optimizer, etc.) necessary to understand the results?
    \item[] Answer:\answerYes{} 
    \item[] Justification: The experimental setting is clearly explained, including all necessary training and test details to understand the results.
    \item[] Guidelines:
    \begin{itemize}
        \item The answer NA means that the paper does not include experiments.
        \item The experimental setting should be presented in the core of the paper to a level of detail that is necessary to appreciate the results and make sense of them.
        \item The full details can be provided either with the code, in appendix, or as supplemental material.
    \end{itemize}

\item {\bf Experiment statistical significance}
    \item[] Question: Does the paper report error bars suitably and correctly defined or other appropriate information about the statistical significance of the experiments?
    \item[] Answer: \answerYes{} 
    \item[] Justification: The paper reports results as averages across scenes and independent runs, following standard practices in the field and providing appropriate statistical reliability.
    \item[] Guidelines:
    \begin{itemize}
        \item The answer NA means that the paper does not include experiments.
        \item The authors should answer "Yes" if the results are accompanied by error bars, confidence intervals, or statistical significance tests, at least for the experiments that support the main claims of the paper.
        \item The factors of variability that the error bars are capturing should be clearly stated (for example, train/test split, initialization, random drawing of some parameter, or overall run with given experimental conditions).
        \item The method for calculating the error bars should be explained (closed form formula, call to a library function, bootstrap, etc.)
        \item The assumptions made should be given (e.g., Normally distributed errors).
        \item It should be clear whether the error bar is the standard deviation or the standard error of the mean.
        \item It is OK to report 1-sigma error bars, but one should state it. The authors should preferably report a 2-sigma error bar than state that they have a 96\% CI, if the hypothesis of Normality of errors is not verified.
        \item For asymmetric distributions, the authors should be careful not to show in tables or figures symmetric error bars that would yield results that are out of range (e.g. negative error rates).
        \item If error bars are reported in tables or plots, The authors should explain in the text how they were calculated and reference the corresponding figures or tables in the text.
    \end{itemize}

\item {\bf Experiments compute resources}
    \item[] Question: For each experiment, does the paper provide sufficient information on the computer resources (type of compute workers, memory, time of execution) needed to reproduce the experiments?
    \item[] Answer: \answerYes{} 
    \item[] Justification: The computational resources are discussed in the Implementation section, providing sufficient information to reproduce the experiments.
    \item[] Guidelines:
    \begin{itemize}
        \item The answer NA means that the paper does not include experiments.
        \item The paper should indicate the type of compute workers CPU or GPU, internal cluster, or cloud provider, including relevant memory and storage.
        \item The paper should provide the amount of compute required for each of the individual experimental runs as well as estimate the total compute. 
        \item The paper should disclose whether the full research project required more compute than the experiments reported in the paper (e.g., preliminary or failed experiments that didn't make it into the paper). 
    \end{itemize}
    
\item {\bf Code of ethics}
    \item[] Question: Does the research conducted in the paper conform, in every respect, with the NeurIPS Code of Ethics \url{https://neurips.cc/public/EthicsGuidelines}?
    \item[] Answer: \answerYes{} 
    \item[] Justification: The research fully conforms with the NeurIPS Code of Ethics in all respects.
    \item[] Guidelines:
    \begin{itemize}
        \item The answer NA means that the authors have not reviewed the NeurIPS Code of Ethics.
        \item If the authors answer No, they should explain the special circumstances that require a deviation from the Code of Ethics.
        \item The authors should make sure to preserve anonymity (e.g., if there is a special consideration due to laws or regulations in their jurisdiction).
    \end{itemize}

\item {\bf Broader impacts}
    \item[] Question: Does the paper discuss both potential positive societal impacts and negative societal impacts of the work performed?
    \item[] Answer: \answerYes{} 
    \item[] Justification: The discussion section addresses both potential societal impacts of the work.
    \item[] Guidelines:
    \begin{itemize}
        \item The answer NA means that there is no societal impact of the work performed.
        \item If the authors answer NA or No, they should explain why their work has no societal impact or why the paper does not address societal impact.
        \item Examples of negative societal impacts include potential malicious or unintended uses (e.g., disinformation, generating fake profiles, surveillance), fairness considerations (e.g., deployment of technologies that could make decisions that unfairly impact specific groups), privacy considerations, and security considerations.
        \item The conference expects that many papers will be foundational research and not tied to particular applications, let alone deployments. However, if there is a direct path to any negative applications, the authors should point it out. For example, it is legitimate to point out that an improvement in the quality of generative models could be used to generate deepfakes for disinformation. On the other hand, it is not needed to point out that a generic algorithm for optimizing neural networks could enable people to train models that generate Deepfakes faster.
        \item The authors should consider possible harms that could arise when the technology is being used as intended and functioning correctly, harms that could arise when the technology is being used as intended but gives incorrect results, and harms following from (intentional or unintentional) misuse of the technology.
        \item If there are negative societal impacts, the authors could also discuss possible mitigation strategies (e.g., gated release of models, providing defenses in addition to attacks, mechanisms for monitoring misuse, mechanisms to monitor how a system learns from feedback over time, improving the efficiency and accessibility of ML).
    \end{itemize}
    
\item {\bf Safeguards}
    \item[] Question: Does the paper describe safeguards that have been put in place for responsible release of data or models that have a high risk for misuse (e.g., pretrained language models, image generators, or scraped datasets)?
    \item[] Answer: \answerNA{} 
    \item[] Justification: The paper poses no such risks.
    \item[] Guidelines:
    \begin{itemize}
        \item The answer NA means that the paper poses no such risks.
        \item Released models that have a high risk for misuse or dual-use should be released with necessary safeguards to allow for controlled use of the model, for example by requiring that users adhere to usage guidelines or restrictions to access the model or implementing safety filters. 
        \item Datasets that have been scraped from the Internet could pose safety risks. The authors should describe how they avoided releasing unsafe images.
        \item We recognize that providing effective safeguards is challenging, and many papers do not require this, but we encourage authors to take this into account and make a best faith effort.
    \end{itemize}

\item {\bf Licenses for existing assets}
    \item[] Question: Are the creators or original owners of assets (e.g., code, data, models), used in the paper, properly credited and are the license and terms of use explicitly mentioned and properly respected?
    \item[] Answer: \answerYes{} 
    \item[] Justification: All datasets and models used in the paper are properly cited, with licenses and terms of use respected.
    \item[] Guidelines:
    \begin{itemize}
        \item The answer NA means that the paper does not use existing assets.
        \item The authors should cite the original paper that produced the code package or dataset.
        \item The authors should state which version of the asset is used and, if possible, include a URL.
        \item The name of the license (e.g., CC-BY 4.0) should be included for each asset.
        \item For scraped data from a particular source (e.g., website), the copyright and terms of service of that source should be provided.
        \item If assets are released, the license, copyright information, and terms of use in the package should be provided. For popular datasets, \url{paperswithcode.com/datasets} has curated licenses for some datasets. Their licensing guide can help determine the license of a dataset.
        \item For existing datasets that are re-packaged, both the original license and the license of the derived asset (if it has changed) should be provided.
        \item If this information is not available online, the authors are encouraged to reach out to the asset's creators.
    \end{itemize}

\item {\bf New assets}
    \item[] Question: Are new assets introduced in the paper well documented and is the documentation provided alongside the assets?
    \item[] Answer: \answerYes{} 
    \item[] Justification: The code is provided in the anonymous URL and includes accompanying documentation.
    \item[] Guidelines:
    \begin{itemize}
        \item The answer NA means that the paper does not release new assets.
        \item Researchers should communicate the details of the dataset/code/model as part of their submissions via structured templates. This includes details about training, license, limitations, etc. 
        \item The paper should discuss whether and how consent was obtained from people whose asset is used.
        \item At submission time, remember to anonymize your assets (if applicable). You can either create an anonymized URL or include an anonymized zip file.
    \end{itemize}

\item {\bf Crowdsourcing and research with human subjects}
    \item[] Question: For crowdsourcing experiments and research with human subjects, does the paper include the full text of instructions given to participants and screenshots, if applicable, as well as details about compensation (if any)? 
    \item[] Answer: \answerNA{} 
    \item[] Justification: This work does not involve crowdsourcing nor research with human subjects.
    \item[] Guidelines:
    \begin{itemize}
        \item The answer NA means that the paper does not involve crowdsourcing nor research with human subjects.
        \item Including this information in the supplemental material is fine, but if the main contribution of the paper involves human subjects, then as much detail as possible should be included in the main paper. 
        \item According to the NeurIPS Code of Ethics, workers involved in data collection, curation, or other labor should be paid at least the minimum wage in the country of the data collector. 
    \end{itemize}

\item {\bf Institutional review board (IRB) approvals or equivalent for research with human subjects}
    \item[] Question: Does the paper describe potential risks incurred by study participants, whether such risks were disclosed to the subjects, and whether Institutional Review Board (IRB) approvals (or an equivalent approval/review based on the requirements of your country or institution) were obtained?
    \item[] Answer: \answerNA{} 
    \item[] Justification: The work does not involve crowdsourcing nor research with human subjects.
    \item[] Guidelines:
    \begin{itemize}
        \item The answer NA means that the paper does not involve crowdsourcing nor research with human subjects.
        \item Depending on the country in which research is conducted, IRB approval (or equivalent) may be required for any human subjects research. If you obtained IRB approval, you should clearly state this in the paper. 
        \item We recognize that the procedures for this may vary significantly between institutions and locations, and we expect authors to adhere to the NeurIPS Code of Ethics and the guidelines for their institution. 
        \item For initial submissions, do not include any information that would break anonymity (if applicable), such as the institution conducting the review.
    \end{itemize}

\item {\bf Declaration of LLM usage}
    \item[] Question: Does the paper describe the usage of LLMs if it is an important, original, or non-standard component of the core methods in this research? Note that if the LLM is used only for writing, editing, or formatting purposes and does not impact the core methodology, scientific rigorousness, or originality of the research, declaration is not required.
    \item[] Answer: \answerNA{} 
    \item[] Justification: The core method development in this research does not involve LLMs.
    \item[] Guidelines:
    \begin{itemize}
        \item The answer NA means that the core method development in this research does not involve LLMs as any important, original, or non-standard components.
        \item Please refer to our LLM policy (\url{https://neurips.cc/Conferences/2025/LLM}) for what should or should not be described.
    \end{itemize}

\end{enumerate}

\end{document}

%% file: preamble.tex
\newcommand{\mtd}{EGGS}
\newcommand{\method}{EGGS\xspace}

%% file: sec/0_abstract.tex
\begin{abstract}

Novel view synthesis (NVS) is crucial in computer vision and graphics, with wide applications in AR, VR, and autonomous driving. While 3D Gaussian Splatting (3DGS) enables real-time rendering with high appearance fidelity, it suffers from multi-view inconsistencies, limiting geometric accuracy. In contrast, 2D Gaussian Splatting (2DGS) enforces multi-view consistency but compromises texture details. To address these limitations, we propose Exchangeable Gaussian Splatting (EGGS), a hybrid representation that integrates 2D and 3D Gaussians to balance appearance and geometry. To achieve this, we introduce {Hybrid Gaussian Rasterization} for unified rendering, {Adaptive Type Exchange} for dynamic adaptation between 2D and 3D Gaussians, and {Frequency-Decoupled Optimization} that effectively exploits the strengths of each type of Gaussian representation. Our CUDA-accelerated implementation ensures efficient training and inference. Extensive experiments demonstrate that EGGS outperforms existing methods in rendering quality, geometric accuracy, and efficiency, providing a practical solution for high-quality NVS. Code and demo available at \href{https://github.com/Fobow/EGGS}{\textcolor{mypink}{https://github.com/Fobow/EGGS}}.
\end{abstract}


%% file: sec/1_intro.tex
\section{Introduction}
\label{sec:intro}

Novel view synthesis (NVS) is a fundamental task in computer graphics and computer vision, with broad applications in augmented reality (AR), virtual reality (VR), and autonomous driving~\cite{chen2024text, li2024art3d, wang2024dc}. Neural Radiance Fields (NeRF)~\cite{mildenhall2021nerf} reconstruct implicit radiance fields via differentiable volume rendering. Despite achieving photorealistic appearance and accurate geometry, NeRF-based methods~\cite{chen2022nerfeff1, tancik2022nerfeff, fridovich2022nerfeff2, kulhanek2023nerfa0, rematas2022nerfa1, tancik2023nerfa2} typically suffer from long training times and slow rendering speeds. 3D Gaussian Splatting (3DGS)~\cite{3dgs} has emerged as an efficient alternative, leveraging anisotropic 3D Gaussians for real-time, high-quality rendering. While 3DGS excels in appearance fidelity, its anisotropic nature often leads to multi-view inconsistencies, limiting geometric accuracy~\cite{2dgs,cheng2024gaussianpro}. As shown in Figure \ref{fig:teaser}, this can lead to inaccurate edges and surfaces.

\begin{figure}[t]
    \centering
    \captionsetup{skip=0.8pt}
    \begin{minipage}{0.3\textwidth}
        \centering
        \includegraphics[width=1\linewidth]{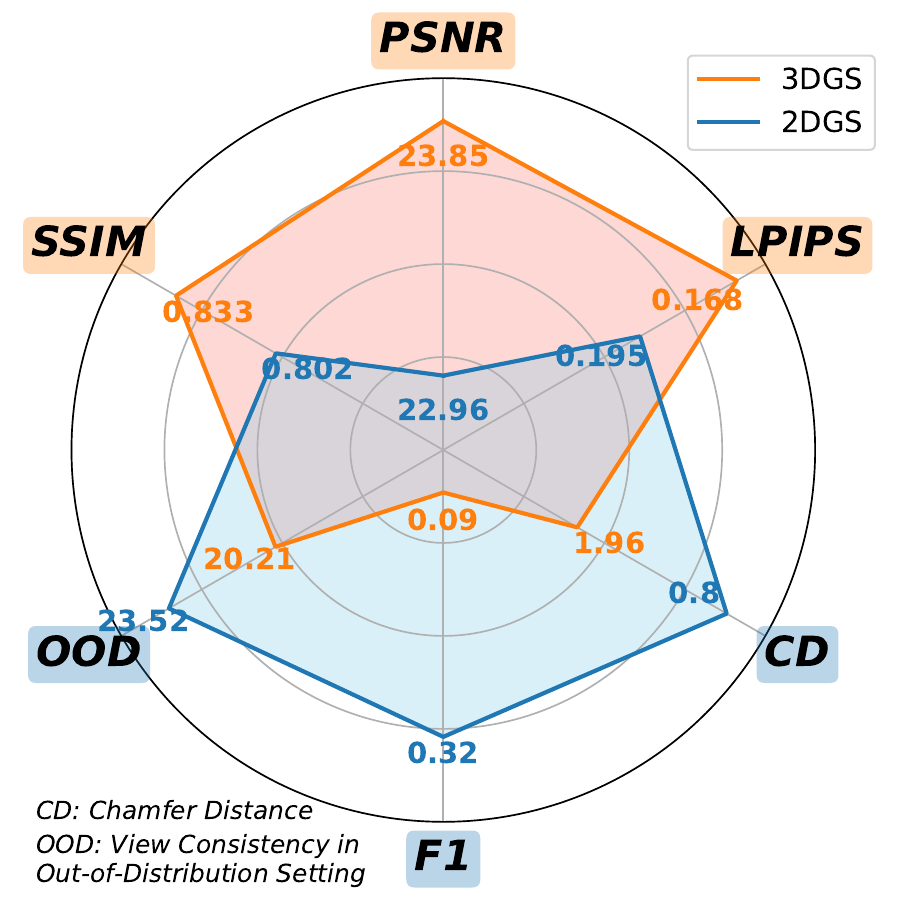}
        \label{fig:motiv}
    \vspace{-0.1in}
    \end{minipage}\hfill
    \begin{minipage}{0.7\textwidth}
        \centering
        \scriptsize
        \setlength{\tabcolsep}{2pt}
        \begin{tabular}{l c c c c c}
        \toprule
        \multirow{2}{*}{\textbf{Method}} 
         & \multirow{2}{*}{\bf \shortstack[c]{Gaussian\\Type}} & \multirow{2}{*}{\bf \shortstack[c]{Raster-\\izer}} & \multirow{2}{*}{\bf \shortstack[c]{Type \\ Exchange}} & \multirow{2}{*}{\bf \shortstack[c]{Regular-\\ization}}  & \multirow{2}{*}{\bf Setting}\\
         & \\
        \midrule
        \textbf{3DGS} \venueTT{SIGGRAPH'23}~\cite{3dgs} &3D    &3D   &\xmark & -  & General   \\
        \textbf{SuGaR} \venueTT{CVPR'24}~\cite{guedon2024sugar} &3D    &3D   &\xmark &Normal  & General     \\
        \textbf{GaussianPro} \venueTT{ICML'24}~\cite{cheng2024gaussianpro} &3D    &3D   &\xmark &Normal   & General  \\
        \midrule
        \textbf{2DGS} \venueTT{SIGGRAPH'24}~\cite{2dgs} &2D    &2D  &\xmark  & Depth \& Normal   & General   \\
        \textbf{GS Surfels} \venueTT{SIGGRAPH'24}~\cite{dai2024surfles} &2D    &3D$^\star$    &\xmark  & Depth \& Normal & General   \\
        \textbf{TextureGS} \venueTT{ECCV'24}~\cite{xu2024texturegs} &2D    &2D   &\xmark & Depth \& Normal   & General   \\
        \midrule
        \textbf{HybridGS} \venueTT{CVPR'25}~\cite{lin2024hybridgs} &3D + 2D*    &3D  &\xmark &-  & Transient  \\
        \textbf{HorizonGS} \venueTT{CVPR'25}~\cite{jiang2024horizon} &3D / 2D    &3D / 2D  &\xmark &Depth \& Normal  & Varying-altitude \\
        \midrule
        \textbf{Ours}   &3D + 2D  &Hybrid &\cmark  &Frequency & General    \\
        \bottomrule
        \end{tabular}
        \label{tab:prior}
    \end{minipage}\hfill
    \caption{Left: Comparison of 3DGS and 2DGS in \colorbox{orange!20}{\strut\textbf{appearance}} and \colorbox{cyan!20}{\strut\textbf{geometry}} metrics. Right: Comparison between \mtd ~and related works. Prior works either use only single representation or do not explore complementary advantages of 3D and 2D Gaussians. $^\star$ Gaussian Surfel~\cite{dai2024surfles} directly sets the $z$-scale of 3D Gaussian to zero and uses the rasterizer from 3DGS. * HybridGS~\cite{lin2024hybridgs} uses image-frame single-view 2D Gaussians~\cite{zhang2024image, zhang2024gaussianimage} instead of 2D Gaussians in the 3D space~\cite{2dgs}.}
    \vspace{-0.25in}
    \label{fig:motiv_merge}
\end{figure}
Following 3DGS, a line of work has focused on improving its geometric accuracy and reconstruction quality through additional regularization and novel representations, as shown in Figure \ref{fig:motiv_merge} (right). SUGAR~\cite{guedon2024sugar} and GaussianPro~\cite{cheng2024gaussianpro} introduce normal-based regularization, such as planar loss, to align Gaussian normals and encourage flatter shapes, thereby improving surface consistency. Gaussian Surfles~\cite{dai2024surfles} and GOF~\cite{yu2024gaussian} incorporate additional geometry-aware constraints to enhance spatial coherence. 2D Gaussian Splatting (2DGS)\cite{2dgs} replaces 3D ellipsoids with 2D surfels, significantly improving multi-view consistency and geometric accuracy, as shown in Figure \ref{fig:motiv_merge} (left). However, this comes at the cost of degraded appearance quality, as surfel-based representations struggle to preserve high-frequency details. TextureGS~\cite{xu2024texturegs} attempts to decouple appearance and geometry within the 2DGS framework, but the single representation still limits overall rendering performance. Recently, HybridGS~\cite{lin2024hybridgs} combines 3DGS with image-space 2D Gaussians to address transient objects, but its radiance field remains fully represented by 3D Gaussians. HorizonGS~\cite{jiang2024horizon}, designed for varying-altitude scenes, decodes 2D Gaussians for surface reconstruction and 3D Gaussians for view synthesis separately via an MLP in ScaffoldGS~\cite{lu2024scaffold}. While effective in their target domains, these methods do not explore a unified hybrid radiance representation. As a result, the complementary strengths of 2DGS and 3DGS in geometry and appearance remain underutilized.

Effectively combining 3D and 2D Gaussians to jointly improve appearance and geometry is non-trivial, as simply mixing the two representations does not necessarily improve reconstruction quality~\cite{jiang2024horizon}. To start, the geometric accuracy of 2D Gaussians relies on a ray–splat–intersection-based rasterizer designed to enforce multi-view consistency. Using the projection-based 3DGS rasterizer to render 2D Gaussians can lead to suboptimal geometry~\cite{dai2024surfles}. Moreover, Gaussian parameters change significantly during training. For instance, 3D Gaussians may flatten to approximate surfaces, while 2D Gaussians may expand volumetrically to capture thin structures or translucent effects. Fixing the Gaussian type throughout optimization can limit the model’s expressiveness. Finally, relying solely on photometric loss is insufficient to balance geometry and appearance. Additional regularization is required to guide the optimization of hybrid representations. Most importantly, the regularization strategy should account for the distinct characteristics of 3D and 2D Gaussians.


In response to these challenges, we introduce \textbf{E}xchan\textbf{g}eable \textbf{G}aussian \textbf{S}platting (EGGS), an adaptive hybrid representation that unifies 2D and 3D Gaussian splatting in a single framework. EGGS provides a practical and efficient solution for high-quality novel view synthesis and 3D reconstruction. Our main contributions are as follows:
\begin{itemize}[leftmargin=15pt,topsep=1pt, itemsep=0.5pt]
    \item To preserve the complementary strengths of 3D and 2D Gaussians, we develop \textit{Hybrid Gaussian Rasterization}, a unified rendering framework that supports both projection-based and ray–splat–intersection-based rasterization. We implement this framework with CUDA for efficient optimization, and ensure compatibility with existing 3DGS and 2DGS pipelines.

    \item We propose \textit{Adaptive Type Exchange}, which enables an exchangeable hybrid of 2D and 3D Gaussians. We use effective rank as an auxiliary criterion to determine whether each Gaussian should dynamically switch its type during training, resulting in a more flexible and content-adaptive representation.

    \item To better balance geometry and appearance, we introduce \textit{Frequency-Decoupled Optimization}, a regularization strategy in the frequency domain. Using the Discrete Wavelet Transform (DWT), we extract low-frequency components to guide scene geometry and high-frequency components to refine appearance. We supervise 3D and 2D Gaussians asymmetrically to exploit their distinct characteristics, where high-frequency signals guide 3D Gaussians toward detailed appearance, while low-frequency signals supervise 2D Gaussians for geometric consistency.

    \item We conduct extensive experiments demonstrating that EGGS significantly improves the trade-off between appearance fidelity and geometric accuracy. It outperforms both 3DGS and 2DGS in appearance quality, while achieving geometric accuracy and multi-view consistency comparable to 2DGS. Moreover, EGGS serves as a versatile representation that performs well in challenging scenarios such as few-shot and out-of-distribution view synthesis.
\end{itemize}

%% file: sec/2_background.tex
\section{Related Works}
\label{sec:related}


\noindent\textbf{Radiance Fields for Novel View Synthesis.} Neural Radiance Fields (NeRF)~\cite{mildenhall2021nerf} have emerged as a fundamental approach for novel view synthesis~\cite{avidan1997nvs}, representing scenes as continuous volumetric functions optimized via differentiable rendering. While NeRF achieves high-fidelity reconstruction, it requires dense sampling and significant computational resources. Subsequent works have improved either quality~\cite{barron2021mip0, barron2022mip} or efficiency~\cite{chen2022nerfeff1, fridovich2022nerfeff2, muller2022ingp, tancik2022nerfeff}, but the excessive training and rendering time remains a major bottleneck. To address this, recent efforts have explored more efficient alternatives, such as 3D Gaussian Splatting (3DGS)~\cite{3dgs}, which represents scenes using a set of 3D Gaussians that can be efficiently rasterized and optimized for real-time rendering. To further improve the performance and efficiency of 3DGS, several extensions have been proposed. ScaffoldGS~\cite{lu2024scaffold} introduces a voxel-based representation where an MLP is used to decode 3D Gaussians within each voxel. 3DGS-MCMC~\cite{kheradmand2024mcmc} formulates Gaussian densification as a Markov Chain Monte Carlo sampling process, enabling a more efficient and adaptive distribution of Gaussians across the scene.

\noindent\textbf{Geometry-Appearance-Balanced Gaussian Splatting.} While 3DGS achieves high appearance fidelity and is efficient in both training and rendering, the anisotropic nature of 3D Gaussians often exhibits multi-view inconsistency, resulting in limited geometric accuracy. To address this, several works propose geometry regularization techniques. SUGAR~\cite{guedon2024sugar} and GaussianPro~\cite{cheng2024gaussianpro} introduce normal-based regularization to encourage flatter Gaussians that better align with scene surfaces. Gaussian Surfels~\cite{dai2024surfles} and GOF~\cite{yu2024gof} further enforce depth accuracy and normal consistency to enhance geometric reconstruction. Instead of relying solely on regularization, 2DGS~\cite{2dgs} adopts a 2D surfel representation with a specialized ray–splat–intersection rasterizer, ensuring multi-view consistency and significantly improving geometric accuracy compared to 3DGS. It also incorporates additional depth and normal regularization. However, this comes at the cost of reduced appearance quality, as 2D surfels struggle to preserve high-frequency detail. TextureGS~\cite{xu2024texturegs} attempts to decouple geometry and appearance modeling within the 2DGS framework, but its appearance fidelity remains limited due to the inherent drawbacks of the 2D representation.

As demonstrated in Figure~\ref{fig:motiv_merge} (left), 3D Gaussians achieve better appearance quality in PSNR, SSIM, and LPIPS. In contrast, 2D Gaussians offer superior view consistency and geometric fidelity, resulting in more robust PSNR under out-of-distribution (OOD) conditions, improved point cloud accuracy in Chamfer Distance (CD), and higher depth accuracy in F1 score. As shown in Figure~\ref{fig:motiv_merge} (right), most existing methods~\cite{2dgs,3dgs,guedon2024sugar,cheng2024gaussianpro,kerbl2024gs0,yu2024mipgs,lu2024scaffold,liu2024maskgaussian,xu2024supergaussians,fan2024lightgaussian,jiang2024gaussianshader} rely on a single Gaussian representation to reconstruct radiance fields, which limits their flexibility and adaptability. Although HybridGS~\cite{lin2024hybridgs} incorporates both 3D Gaussians and image-space 2D Gaussians to better handle transient content, its radiance field remains solely represented by 3D Gaussians. A radiance field that jointly leverages both 2D and 3D Gaussians remains largely unexplored. It is still unclear how 2D and 3D Gaussians can be made exchangeable during training and how to fully exploit their complementary strengths in appearance and geometry. We provide a more detailed discussion in Appendix~\ref{app:related}.

%% file: sec/3_method.tex
\section{Method}
\label{sec:method}

\begin{figure*}[t]
    \centering
    \includegraphics[width=0.95\linewidth]{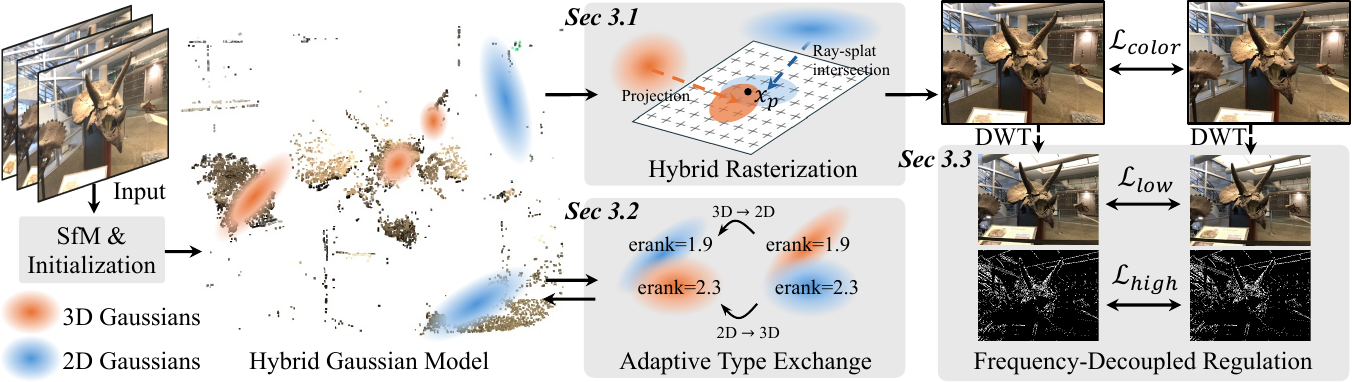}
    \caption{Overview of the \method framework. We initialize 2D and 3D Gaussians from sparse points obtained via structure-from-motion (SfM)~\cite{schonberger2016colmap, schonberger2016colmap2}. Their parameters are then jointly optimized using our CUDA-accelerated differentiable hybrid rasterization. To enhance the flexibility of the hybrid representation, Adaptive Type Exchange is introduced to allow each Gaussian to switch between 2D and 3D types during training. Finally, we apply Discrete Wavelet Transform (DWT)~\cite{heil1989dwt} and introduce Frequency-Decoupled Optimization to balance geometric accuracy and appearance fidelity.}
    \label{fig:overview}
    \vspace{-0.2in}
\end{figure*}

We provide an overview of the \method framework in Figure~\ref{fig:overview}. To enable the joint training of 2D and 3D Gaussians within a unified framework, we first introduce \textit{Hybrid Gaussian Rasterization} in Section~\ref{sec:hyras}, which supports both ray–splat–intersection-based rendering for 2D Gaussians and projection-based rendering for 3D Gaussians. Next, we present \textit{Adaptive Type Exchange} in Section~\ref{sec:ats}, which enables dynamic switching between 2D and 3D types during optimization. Finally, to optimize the hybrid model for balanced geometric consistency and appearance fidelity, we propose \textit{Frequency-Decoupled Optimization} in Section~\ref{sec:wave}, a supervision strategy that leverages the distinct frequency characteristics of 2D and 3D Gaussians.


\subsection{Hybrid Gaussian Rasterization}
\label{sec:hyras}
Differentiable rasterization was introduced in 3DGS to enable gradient-based optimization of Gaussian parameters using a projection-based pipeline for real-time rendering. 2DGS later developed a ray–splat–intersection-based rasterizer tailored to 2D surfel representations, improving multi-view consistency and geometric accuracy. However, the architectural distinction between these two rasterization pipelines makes it non-trivial to render and optimize a hybrid model within a unified framework. While 2D Gaussians can be viewed as degenerate 3D Gaussians with zero scale along the $z$-axis, directly rendering them with the 3D rasterizer leads to geometric inaccuracies~\cite{2dgs}. This is due to the affine projection approximation used in 3DGS, which introduces distortion at all points except the Gaussian center. We further analyze this issue in Section~\ref{sec:ablation}.




To leverage the complementary strengths of 3D and 2D Gaussians, it is necessary to render them within a unified framework. To this end, we propose \textit{Hybrid Gaussian Rasterization}, which integrates both projection-based and ray–splat–intersection-based pipelines. In our rasterizer, each Gaussian primitive $\mathcal{G}$ is parameterized by a center $\boldsymbol{\mu} \in \mathbb{R}^3$, scale $\boldsymbol{s} \in \mathbb{R}^3$, rotation quaternion $\boldsymbol{r} \in \mathbb{R}^4$, opacity $\alpha \in \mathbb{R}$, and spherical harmonic (SH) color coefficients $\boldsymbol{f} \in \mathbb{R}^{3 \times (l+1)^2}$, where $l$ is the degree of view-dependent color. The view-dependent RGB color $\boldsymbol{c}$ is decoded from $\boldsymbol{f}$. The Gaussian shape is defined by the covariance matrix $\boldsymbol{\Sigma} = \boldsymbol{R} \boldsymbol{S} \boldsymbol{S}^T \boldsymbol{R}^T$, where $\boldsymbol{R} \in \mathbb{R}^{3 \times 3}$ is the rotation matrix derived from $\boldsymbol{r}$, and $\boldsymbol{S} = \text{diag}(s_x, s_y, s_z) \in \mathbb{R}^{3 \times 3}$ is the scaling matrix. We augment each Gaussian with a type specifier $t \in \{0, 1\}$ to indicate whether it is a 2D ($t = 0$) or 3D ($t = 1$) Gaussian. 2D Gaussians are initialized with $s_z = 0$, while the remaining parameters follow the initialization of 3DGS.

\begin{wrapfigure}{r}{0.5\textwidth}
    \vspace{-0.1in}
    \centering
    \includegraphics[width=0.8\linewidth]{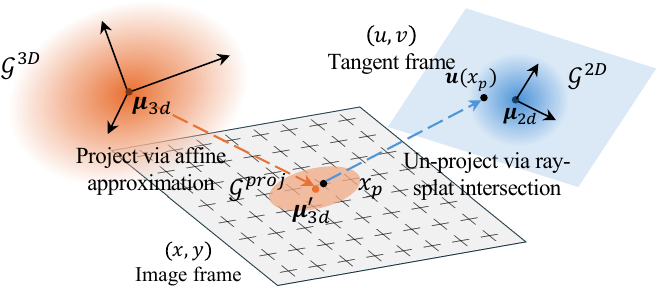}
    \caption{Illustration of Hybrid Gaussian Rasterization. The contribution of 3D Gaussians and 2D Gaussians is computed via affine projection and ray-splat-intersection, respectively.}
    \label{fig:hyras}
    \vspace{-0.22in}
\end{wrapfigure}
As shown in Figure~\ref{fig:hyras}, we rasterize Gaussians according to their types, where affine projection is used for 3D Gaussians and ray–splat–intersection is used for 2D Gaussians. The contribution of $\mathcal{G}^{3d}_i$ and $\mathcal{G}^{2d}_i$ is evaluated by computing the distance from the image-space pixel $x_p$ to $\mathcal{G}^{proj}_i$ or $\mathcal{G}^{2d}_i$ in the 2D image plane or tangent frame, respectively:
\begin{equation}
\scriptsize
\label{e:dist2}
    d_i = \begin{cases}
    (\boldsymbol{u}_i(x_p)^2+\boldsymbol{v}_i(x_p)^2
     & \text{if } t_i=0 \\
       (x_p-\boldsymbol{\mu}'_{3d,i})^T\boldsymbol{\Sigma}'^{-1}_i(x_p - \boldsymbol{\mu}'_{3d,i}) & \text{otherwise}
    \end{cases}
\end{equation}
where $\boldsymbol{\mu}'_{3d,i}$ and $\boldsymbol{\Sigma}_i$ are the projected center and covariance of the 3D Gaussian computed via affine projection, and $u_i(x_p)$ and $v_i(x_p)$ denote the coordinates of the intersection between the ray through $x_p$ and the 2D Gaussian. The distance $d_i$ can be computed simultaneously for both 3D and 2D Gaussians. The final contribution of each Gaussian is then computed uniformly as $\tilde{\alpha}_i = \alpha_i e^{-\frac{1}{2}d_i}$, where $\alpha_i$ is the opacity of the $i$-th Gaussian. With the above formulation, both 3D and 2D Gaussians can be rendered in a single $\alpha$-blending pass:
\begin{equation}
\label{e:blend}
    C(x_p) = \textstyle\sum_{i \in N} \boldsymbol{c}_i \tilde{\alpha}_i \prod_{j=1}^{i-1} (1 - \tilde{\alpha}_j)
\end{equation}
where the final color at pixel $x_p$ is computed from color $\boldsymbol{c}_i$ and contribution $\tilde{\alpha}_i$ of each Gaussian primitive. To support efficient and parallel rendering, we implement our hybrid rasterizer in CUDA. More details on initialization and densification are provided in Appendix~\ref{app:details}, and those on projection-based and ray–splat–intersection-based rasterization procedures are deferred to Appendix~\ref{app:rasterization}.

\subsection{Adaptive Type Exchange}
\label{sec:ats}
While the type specifier introduced in Section~\ref{sec:hyras} enables unified rendering of 2D and 3D Gaussians, each Gaussian primitive is initialized with a fixed type. Such fixed type assignment can limit the expressiveness of the model, as Gaussians may naturally deviate from their initial type during optimization. For example, 3D Gaussians may become increasingly flat to better model surfaces, while 2D Gaussians may take on more volumetric properties to capture semi-transparent regions. To fully exploit the flexibility of the hybrid model, the type of each Gaussian should dynamically adapt to its evolving geometric characteristics. To this end, we propose \textit{Adaptive Type Exchange}, which allows each Gaussian to switch between 2D and 3D types during training.

\begin{wrapfigure}{r}{0.5\textwidth}
    \centering
    \captionsetup{skip=0.1pt}
    \includegraphics[width=0.9\linewidth]{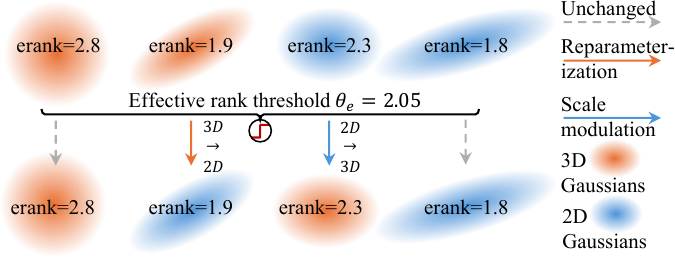}
    \caption{Illustration of Adaptive Type Exchange.}
    \label{fig:ats}
\end{wrapfigure}

The key to \textit{Adaptive Type Exchange} is detecting discrepancies between a Gaussian’s assigned type and its effective geometric dimensionality. Therefore, we introduce the effective rank (erank)~\cite{roy2007erank, hyung2024erankgs} as an indicator of this dimensionality, allowing the model to determine when type switching is needed during training. Given a Gaussian $\mathcal{G}$ with scaling $\boldsymbol{s} = (s_x, s_y, s_z)$, we define its effective rank as:
\begin{equation}
\textstyle
\text{erank}(\mathcal{G}) = \exp\left(-\sum_{i=0}^{2} \frac{q_i}{\|\boldsymbol{q}\|_1} \log \frac{q_i}{\|\boldsymbol{q}\|_1} \right), \quad \text{where} \ \boldsymbol{q} = (s_x^2, s_y^2, s_z^2).
\end{equation}
As illustrated in Figure~\ref{fig:ats}, erank provides a principled signal for deciding when to switch types. A perfectly isotropic 3D Gaussian has $\text{erank} = 3$, while a flattened Gaussian approaches $\text{erank} = 2$. If a Gaussian primitive $\mathcal{G}_i$ is assigned as 3D ($t_i = 1$) but its effective rank falls below a threshold $\theta_e$, we mark it for conversion to 2D by setting $t_i' = 0$. Similarly, we update 2D Gaussians to 3D ($t_i' = 1$) when their effective rank exceeds the threshold. Yet, merely flipping the type specifier can lead to unstable parameter transitions, as the $s_z$ scale is treated as least significant in 2D Gaussians. To ensure stable conversion, we reparameterize the covariance of 3D Gaussians during switching and adjust gradient flow to $s_z$ for 2D Gaussians.




\noindent\textbf{Reparameterization. ($3D\rightarrow2D$)} All three scales of a 3D Gaussian are initially optimized, whereas 2D Gaussians ignore the $s_z$ scale during ray–splat–intersection-based rasterization. Accordingly, when converting a 3D Gaussian to 2D, only $s_x$ and $s_y$ are retained and $s_z$ is discarded. However, directly discarding $s_z$ can lead to instability during training when it is not the least significant scale. To prevent this, we reparameterize the 3D Gaussian before conversion so that $s_z$ corresponds to the smallest axis. The key to stable conversion is aligning $s_z$ with the least significant scale while preserving covariance $\boldsymbol{\Sigma} = \boldsymbol{R} \boldsymbol{S} \boldsymbol{S}^T \boldsymbol{R}^T$.
We first construct the converted scaling matrix $\boldsymbol{S}^*$
using a permutation matrix $\boldsymbol{P}$ that moves
the least significant scale to the $z$-axis:
\begin{equation}
\textstyle
\small
    \boldsymbol{S}^* = \boldsymbol{P}\boldsymbol{SP}^T
\end{equation}
$\boldsymbol{P}$ is set to $\boldsymbol{P}_x$ or $\boldsymbol{P}_y$ if $s_x$ or $s_y$ is the least significant scale, respectively, where:
\begin{equation}
\textstyle
\small
    \boldsymbol{P}_x = \begin{bmatrix}
            0 & 1 & 0 \\
            0 & 0 & 1 \\
            1 & 0 & 0
        \end{bmatrix}, \quad
    \boldsymbol{P}_y = \begin{bmatrix}
            0 & 0 & 1 \\
            1 & 0 & 0 \\
            0 & 1 & 0
        \end{bmatrix}
\end{equation}
Then, to ensure the covariance $\boldsymbol{\Sigma}$ remain unchanged in $\boldsymbol{\Sigma}^* = \boldsymbol{R}^*\boldsymbol{S}^*\boldsymbol{S}^*\boldsymbol{R}^{*T}$, we set the rotation of 2D Gaussian as $\boldsymbol{R}^*=\boldsymbol{RP}^T$. We note that $\boldsymbol{R}^*$ is converted to quaternions during optimization. To ensure a valid conversion, the rotation matrix $\boldsymbol{R}^*$ must be orthogonal with a positive determinant. $\boldsymbol{P}_x$ and $\boldsymbol{P}_y$ are designed to preserve these properties. Additional details are provided in Appendix~\ref{app:repa}.

\noindent\textbf{Scale Modulation. ($2D\rightarrow3D$)}  
When converting 2D Gaussians to 3D, all parameters are retained with the type specifier flipped. However, as mentioned above, the $s_z$ scale of 2D Gaussians is not optimized during rasterization. However, to allow 2D Gaussians to develop volumetric capacity and transition to 3D when needed, gradient flow must also be introduced along the $z$-axis. To support this, we incorporate $s_z$ into the computation graph via a soft modulation based on opacity $\alpha$,
\begin{wrapfigure}{r}{0.5\textwidth}
    \centering
    \includegraphics[width=1\linewidth]{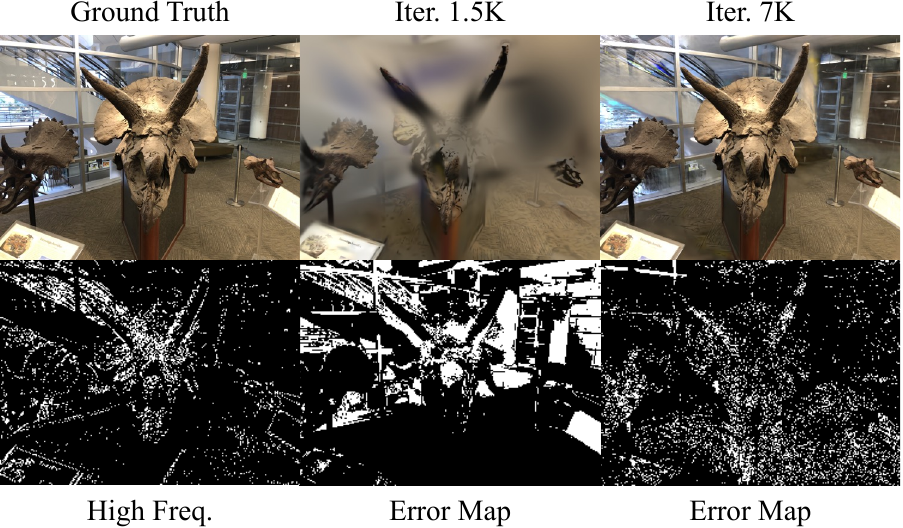}
    \caption{Illustration of reconstruction error during training. In early iterations, the model focuses on overall scene geometry, while high-frequency local details are progressively refined in later stages.}
    \label{fig:dwt}
    \vspace{0.05in}
\end{wrapfigure}
aligning geometric expressiveness with visual transparency. The intuition behind this design is that 2D-to-3D transitions often occur in regions with semi-transparent or volumetric effects that flat primitives cannot represent well. This modulation enables $s_z$ to be optimized throughout training, while ensuring its updates remain stable and smoothly conditioned on opacity:
\begin{equation}
\textstyle
    \alpha_i^* = \alpha_i e^{-\lambda_z\times s_z^{*}}
\end{equation}
where $s_z^*$ denotes the activated $z$-axis scale, computed via a soft gating function:
\begin{equation}
\label{e:soft}
\textstyle
    s_z^{*} = \text{sigmoid}\left(\frac{s_z - \theta_z}{T_z}\right)s_z
\end{equation}
The soft scale modulation allows a 2D Gaussian to remain effectively two-dimensional when $s_z$ is insignificant, in which case $s_z^*$ approaches zero. Conversely, as a 2D Gaussian evolves toward a more volumetric form, an increasing $s_z$ leads to reduced opacity, effectively enabling the representation of semi-transparent or volumetric effects. Additional details on the effective rank threshold, 3D Gaussian reparameterization and permutation, and 2D Gaussian scale modulation are provided in Appendix~\ref{app:repa}.









\subsection{Frequency-Decoupled Optimization}
\label{sec:wave}
With our hybrid rasterization and adaptive type exchange mechanism, 2D and 3D Gaussians can be jointly optimized within a unified and flexible framework. However, relying solely on photometric loss is insufficient to effectively optimize the hybrid model for balanced geometry and appearance. 2D and 3D Gaussians exhibit distinct characteristics during optimization and specialize in different aspects of the scene. 2D Gaussians are better suited for enforcing geometric consistency, while 3D Gaussians excel at capturing high-frequency appearance details. To fully leverage these complementary strengths, we introduce \textit{Frequency-Decoupled Optimization}, a supervision strategy that decouples low- and high-frequency components and assigns them asymmetrically to 2D and 3D Gaussians, respectively.

\textbf{Frequency Decoupling via Discrete Wavelet Transform.}
As shown in Figure~\ref{fig:dwt}, scene information can be effectively separated in the frequency domain. High-frequency components typically correspond to fine details that are refined in later training stages (e.g., 7K iterations), while low-frequency components capture overall scene geometry and are optimized earlier. This frequency-based separation aligns well with the complementary roles of 3D Gaussians in modeling appearance and 2D Gaussians in capturing geometry. Motivated by this, we introduce \textit{Frequency-Decoupled Optimization} to supervise the hybrid model in the frequency domain. We apply DWT~\cite{heil1989dwt} to decompose the ground truth image $\mathcal{I}$ into low- and high-frequency components: $\mathcal{I}_l, \mathcal{I}_h = \text{DWT}(\mathcal{I})$. The same transformation is applied to the rendered image $\hat{\mathcal{I}}$ to obtain $\hat{\mathcal{I}}_l$ and $\hat{\mathcal{I}}_h$.
\begin{wrapfigure}{r}{0.45\textwidth}
    \vspace{-0.1in}
    \centering
    \includegraphics[width=1\linewidth]{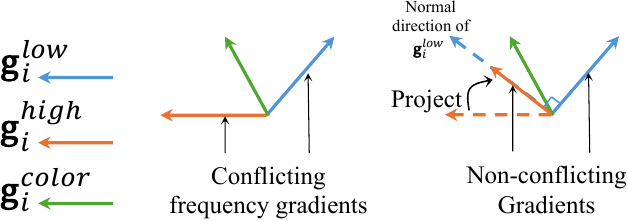}
    \caption{Illustration of frequency gradient projection. We project gradients from high-frequency loss onto the normal vector of gradients from low-frequency for 2D Gaussians.}
    \label{fig:pc}
\end{wrapfigure}
The frequency loss is defined as $\mathcal{L}_i = \|\hat{\mathcal{I}}_i - \mathcal{I}_i\|_2^2$ for $i \in \{\text{low}, \text{high}\}$. We also include the standard appearance loss used in 3DGS: $\mathcal{L}_{\text{color}} = (1 - \lambda) \mathcal{L}_1 + \lambda \mathcal{L}_{\text{D-SSIM}}$. With access to frequency-specific losses, a naïve strategy is to combine all terms and apply them uniformly to all Gaussians:
\begin{equation}
\label{e:loss}
\mathcal{L} = \mathcal{L}_{\text{color}} + \lambda_{\text{low}} \mathcal{L}_{\text{low}} + \lambda_{\text{high}} \mathcal{L}_{\text{high}}.
\end{equation}
where $\mathcal{L}_{\text{low}}$ and $\mathcal{L}_{\text{high}}$ are applied equally to all 2D and 3D Gaussians. While supervision is decoupled in the frequency domain, this approach overlooks the distinctions of each representation.

\textbf{Asymmetrical Gradient Update with Projected Conflicts.} We denote the gradients to $\mathcal{G}_i$ from $\mathcal{L}_{\text{color}}, \mathcal{L}_{\text{low}}$ and $\mathcal{L}_{\text{high}}$ as $\mathbf{g}_i^{color}$, $\mathbf{g}_i^{low}$ and $\mathbf{g}_i^{high}$, respectively. 
As illustrated in Figure~\ref{fig:pc}, conflicting gradients can arise when losses from different frequency components are directly applied to update Gaussian parameters (i.e., Eq.(\ref{e:loss})), diminishing the effectiveness of frequency-based regularization. Such conflicts stem from the distinct characteristics of 2D and 3D Gaussians. 2D Gaussians are more effective at capturing overall geometry and ensuring multi-view consistency, where low-frequency signals offer more relevant guidance, while high-frequency gradients may counteract this by encouraging appearance-driven updates. Conversely, 3D Gaussians
\begin{wrapfigure}{r}{0.55\textwidth}
    \vspace{-0.15in}
    \begin{algorithm}[H]
    \footnotesize
    \caption{Frequency-Decoupled Optimization}
    \Require {Gaussians $\{\mathcal{G}_i\}_{i=0}^{N-1}$, appearance loss $\mathcal{L}_{color}$, frequency loss $\mathcal{L}_{low}$ and $\mathcal{L}_{high}$.}
    $\mathbf{g}^{color}, \mathbf{g}^{low}, \mathbf{g}^{high} \leftarrow \nabla_{\mathcal{G}} \mathcal{L}_{color}, \nabla_{\mathcal{G}} \mathcal{L}_{low} ,\nabla_{\mathcal{G}} \mathcal{L}_{high}$\;
    {\scriptsize \texttt{//} Process per Gaussian gradient conflict} \\
    \For {$i \leftarrow 0$ \KwTo $N - 1$ }{
    {\scriptsize \texttt{//} There are conflict Gradients in different frequencies } \\
        \If{$\mathbf{g}^{low}_i \cdot \mathbf{g}^{high}_i < 0 $}{
            \uIf{$t_i == 0$}{ 
                $\mathbf{g}^{high}_i \leftarrow \mathbf{g}^{high}_i 
                  - \frac{\mathbf{g}^{high} \cdot \mathbf{g}^{low}}{||\mathbf{g}^{low}||^2} 
                  \mathbf{g}^{low}$; \hfill ~(9)\label{e:proj_high} \\ 
                \hfill {\scriptsize \texttt{//} Type is 2D, project $\mathbf{g}^{high}_i$ onto normal of $\mathbf{g}^{low}_i$}
            }
            \Else{
                $\mathbf{g}^{low}_i \leftarrow \mathbf{g}^{low}_i - \frac{\mathbf{g}^{high} \cdot \mathbf{g}^{low}}{||\mathbf{g}^{high}||^2} \mathbf{g}^{high}$; \hfill ~(10)\label{e:proj_low} \\
                \hfill {\scriptsize \texttt{//} Type is 3D, project $\mathbf{g}^{low}_i$ onto normal of $\mathbf{g}^{high}_i$}
            }
        }
        $\Delta \mathcal{G}_i = \mathbf{g}^{color}_{i} +\mathbf{g}^{low}_{i} + \mathbf{g}^{high}_{i}$
    }
    \Return Update $\Delta \mathcal{G}$
    \label{alg:fdo}
    \end{algorithm}
    \vspace{-0.2in}
\end{wrapfigure}
specialize in modeling fine-scale appearance and benefit more from high-frequency supervision, whereas low-frequency signals contribute less to their performance.


To address this issue, we propose an asymmetrical update strategy that applies frequency supervision based on Gaussian type, as shown in Algorithm~\ref{alg:fdo}.
For each Gaussian, we check for potential gradient conflicts by computing the inner product between $\mathbf{g}_i^{\text{low}}$ and $\mathbf{g}_i^{\text{high}}$, where a negative value indicates divergent update directions~\cite{yu2020gradient}. When such conflict is detected, we retain the frequency component most relevant to the Gaussian type and project the other. Specifically, for 2D Gaussians, we preserve supervision from low-frequency and remove the conflicting component of high-frequency by projecting $\mathbf{g}_i^{\text{high}}$ onto the normal vector of $\mathbf{g}_i^{\text{low}}$, as shown in Eq.(9). Similarly, for 3D Gaussians, we retain $\mathbf{g}_i^{\text{high}}$ and project $\mathbf{g}_i^{\text{low}}$ as indicated in Eq.(10). This asymmetrical supervision ensures each Gaussian is updated along its most informative direction while minimizing interference from less relevant frequency signals. More details on DWT and the gradient projection strategy are provided in  Appendix~\ref{app:dwt} and Appendix~\ref{app:proj}, respectively.

%% file: sec/4_expt.tex
\label{sec:expt}
\section{Experiments}
\noindent\textbf{Datasets and Metrics.}
We evaluate \method on several widely used benchmarks. For appearance evaluation, we use Mip-NeRF360~\cite{barron2022mip}, LLFF~\cite{mildenhall2019llff}, Tanks\&Temples~\cite{knapitsch2017tanks}, and DTU~\cite{jensen2014dtu}. For geometry evaluation, we use DTU, which provides ground-truth point clouds, and Tanks\&Temples, which offers ground-truth depth maps. Additional dataset details are provided in Appendix~\ref{app:details}. Following prior work~\cite{barron2022mip, 3dgs, cheng2024gaussianpro, 2dgs}, we report PSNR, SSIM~\cite{wang2004psnr}, and LPIPS~\cite{zhang2018lpips} to evaluate the appearance quality of synthesized novel views. For geometry, we follow~\cite{2dgs, hyung2024erankgs} and report Chamfer Distance~\cite{barrow1978cd} on DTU to assess reconstruction accuracy.

\begin{figure*}[t]
    \vspace{-0.2in}
    \centering
    \includegraphics[width=1\linewidth]{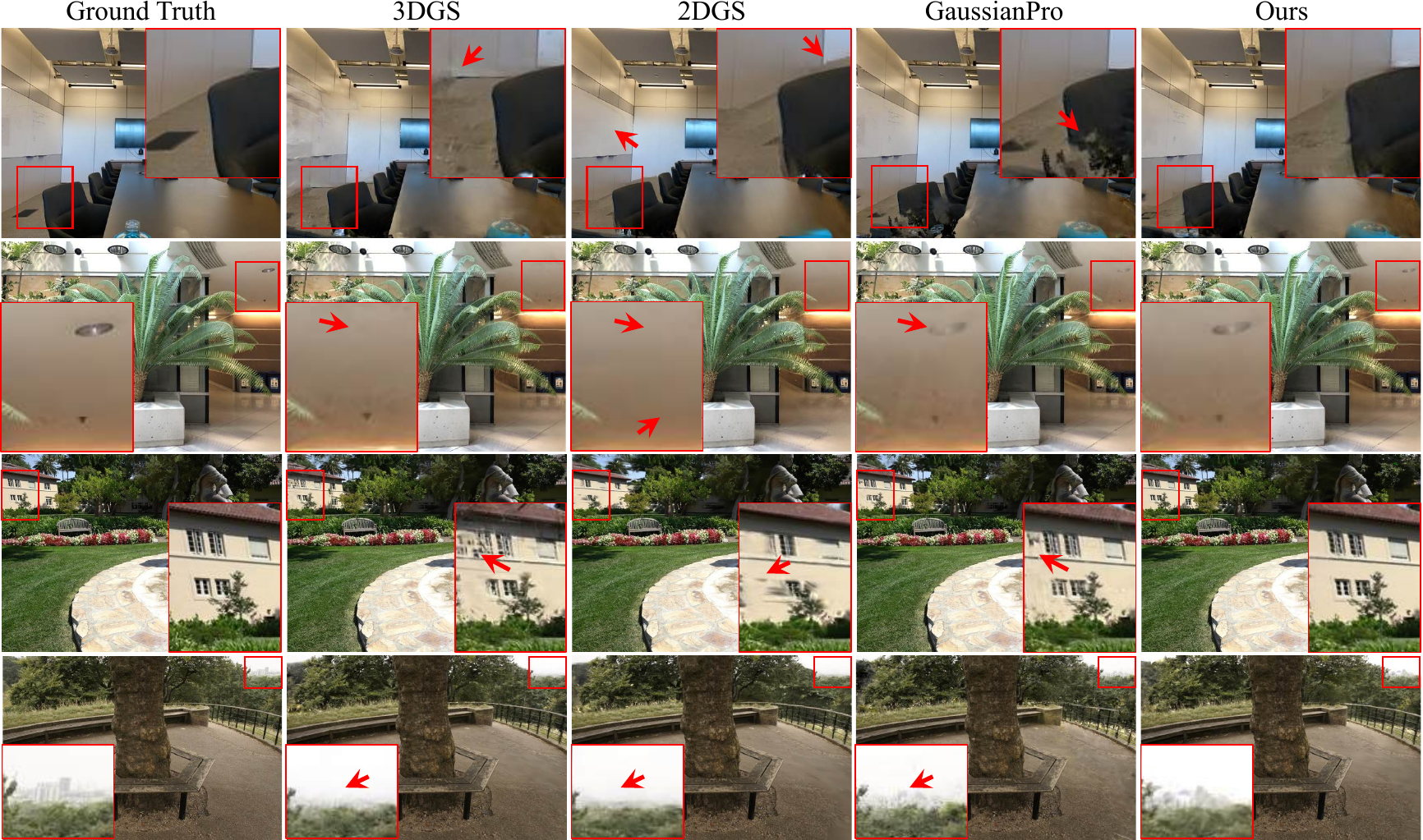}
    \caption{\small Qualitative comparison on LLFF, Tanks\&Temples, and Mip-NeRF360. 3DGS suffers from inaccurate scene geometry. While 2DGS improves geometric fidelity, it overlooks texture and local details. \method recovers more accurate geometry while preserving high-frequency details. Additional visual results and videos are available in the supplementary material and \href{https://github.com/Fobow/EGGS}{\textcolor{mypink}{project website}}.}
    \label{fig:app}
    \vspace{-0.15in}
\end{figure*}

\noindent\textbf{Baselines.}
To demonstrate the effectiveness of \method, we compare against several single-representation methods that use either 3D or 2D Gaussians. For 3D Gaussian-based methods, we include vanilla 3DGS~\cite{3dgs}, GaussianPro~\cite{cheng2024gaussianpro} and GOF~\cite{yu2024gof}, which incorporate geometric regularization, and FreGS~\cite{zhang2024fregs}, which introduces frequency-based supervision. For 2D Gaussian-based methods, we consider vanilla 2DGS~\cite{2dgs} and TextureGS~\cite{xu2024texturegs}, which improves the appearance fidelity of 2D Gaussians. Additional discussion of related methods is provided in Appendix~\ref{app:related}.

\noindent\textbf{Implementation.} We implement our hybrid rasterizer based on the CUDA rasterization code of 3DGS~\cite{3dgs}. We used the Haar filter for the DWT~\cite{heil1989dwt, strang1996wavelets}. For Frequency-Decoupled Optimization, we set the weight for the frequency components as $\lambda_{low} = 0.2$ and $\lambda_{low} = 0.4$. For Adaptive Type Switch, we set the erank threshold as $2.05$. We offer more details about our training pipeline and parameter setting in Appendix~\ref{app:details}. All experiments are conducted on an A5000 GPU.

\subsection{Results}


\begin{table*}[t]
\captionsetup{skip=2pt}
\centering
\scriptsize
\caption{ \small Quantitative comparison on Mip-NeRF360, LLFF and Tanks\&Temples datasets. The best, second-best, and third-best entries are marked in \colorbox{myred}{\strut red}, \colorbox{myorange}{\strut orange}, and \colorbox{myyellow}{\strut yellow}, respectively. }
\setlength{\tabcolsep}{5pt}
\begin{tabular}{l ccc ccc ccc}
\toprule
\multirow{2.5}{*}{\textbf{Method}} 
& \multicolumn{3}{c}{\textbf{Mip-NeRF360}} 
& \multicolumn{3}{c}{\textbf{LLFF}} 
& \multicolumn{3}{c}{\textbf{Tanks\&Temples}} \\
\cmidrule(lr){2-4} \cmidrule(lr){5-7} \cmidrule(lr){8-10}
& PSNR $\uparrow$ & SSIM $\uparrow$ & LPIPS $\downarrow$
& PSNR $\uparrow$ & SSIM $\uparrow$ & LPIPS $\downarrow$
& PSNR $\uparrow$ & SSIM $\uparrow$ & LPIPS $\downarrow$\\
\midrule
2DGS \venueTT{SIGGRAPH'24}~\cite{2dgs} &26.81 &0.796 &0.297 &24.93 &0.815 &0.147 &22.96 &0.802 &0.195 \\
TextureGS \venueTT{ECCV'24} &27.14 &0.803 &0.285 &25.58 &0.837 &0.117 &22.43 &0.811 &0.189 \\
3DGS \venueTT{SIGGRAPH'23}~\cite{3dgs} &27.43 &0.814 &0.257 &\cellcolor{myyellow}26.12 &\cellcolor{myyellow}0.865 &\cellcolor{myorange}0.099 &23.85 &0.833 &\cellcolor{myyellow}0.168 \\
GOF \venueTT{TOG'24}~\cite{yu2024gof} &27.42 &\cellcolor{myorange}0.826 &0.234 &25.57 &0.854 &0.121 &22.41 &0.831 &0.172 \\
GaussianPro \venueTT{ICML'24}~\cite{cheng2024gaussianpro} &\cellcolor{myorange}27.92 &\cellcolor{myyellow}0.825 &\cellcolor{myorange}0.208 &\cellcolor{myorange}26.53 &\cellcolor{myorange}0.867 &0.105 &\cellcolor{myyellow}23.92 &\cellcolor{myorange}0.855 &\cellcolor{myorange}0.162 \\
FreGS \venueTT{CVPR'24}~\cite{zhang2024fregs} &\cellcolor{myyellow}27.85 &\cellcolor{myorange}0.826 &\cellcolor{myyellow}0.209 &26.11 &0.860 &\cellcolor{myyellow}0.102 &\cellcolor{myorange}23.96 &\cellcolor{myyellow}0.849 &0.178 \\
\hline
Ours &\cellcolor{myred}27.96 &\cellcolor{myred}0.851 &\cellcolor{myred}0.192 &\cellcolor{myred}27.34 &\cellcolor{myred}0.895 &\cellcolor{myred}0.083 &\cellcolor{myred}24.41 &\cellcolor{myred}0.923 &\cellcolor{myred}0.153 \\
\hline
\end{tabular}
\vspace{-0.2in}
\label{tab:app}
\end{table*}

\noindent\textbf{Appearance.}
Figure~\ref{fig:app} and Table~\ref{tab:app} show qualitative and quantitative comparisons on Mip-NeRF360, LLFF, and Tanks\&Temples. 3D Gaussian-based methods generally achieve better PSNR, SSIM, and LPIPS but often produce blurred geometry due to anisotropic Gaussians. GaussianPro and FreGS improve reconstruction via geometric or frequency regularization but still lack geometric accuracy. 2DGS produces cleaner edges and better geometry, yet oversmooths details and underperforms in appearance. TextureGS enhances 2D appearance but remains inferior to 3D-based methods. In contrast, \method outperforms all baselines by combining the strengths of 2D and 3D Gaussians. It recovers more accurate geometry while preserving high-frequency visual details.

\begin{figure*}[t]
    \vspace{-0.2in}
    \centering
    \captionsetup{skip=0.8pt}
    \includegraphics[width=0.95\linewidth]{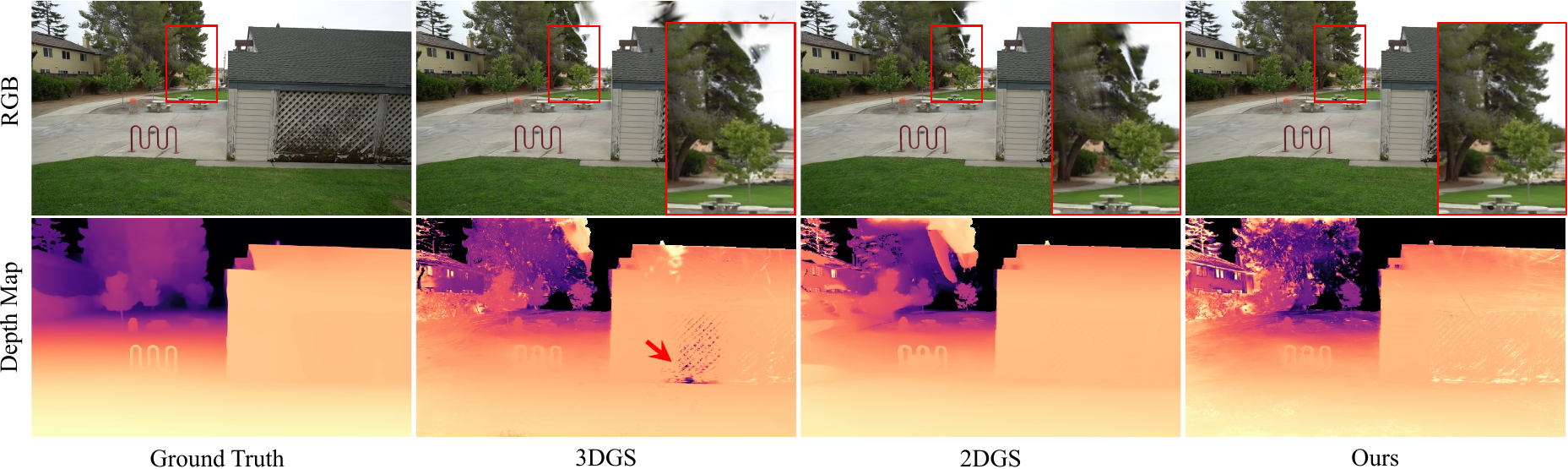}
    \caption{\small Qualitative comparison on Tanks\&Temples. \method achieves better overall reconstruction quality, producing more accurate depth maps than 3DGS and recovering high-frequency details better than 2DGS.}
    \label{fig:tandt}
    \vspace{-0.1in}
\end{figure*}

\noindent\textbf{Geometry.}
We evaluate geometry reconstruction quality on Tanks\&Temples and DTU. As shown in Figure~\ref{fig:tandt}, both 2DGS and \method produce more accurate depth maps than 3DGS, with sharper surfaces and clearer edges. However, 2DGS sacrifices appearance fidelity due to the lack of high-frequency detail. In contrast, \method improves geometry over 3DGS while also preserving appearance quality. Table~\ref{tab:dtucd} reports Chamfer Distance on the DTU dataset, where \method outperforms 3DGS and SUGAR. Note that SUGAR, 2DGS, and GOF prioritize surface reconstruction and mesh extraction, often at the cost of appearance. Although 2DGS is slightly more accurate geometrically, \method achieves a better trade-off, offering stronger appearance quality alongside competitive geometry.

\begin{table*}[b]
    \centering
    \captionsetup{skip=1pt}
    \caption{\small Quantitative Geometry Comparison on DTU.
Chamfer Distance (CD) is reported per scene. mCD and PSNR denote the mean Chamfer Distance and mean PSNR across all scenes, respectively.}
    \label{tab:comparison}
    \resizebox{1\linewidth}{!}{
    \begin{tabular}{lccccccccccccccc|cc}
        \toprule
        Method & 24  & 37  & 40  & 55  & 63  & 65  & 69  & 83  & 97  & 105 & 106 & 110 & 114 & 118 & 122 & mCD & PSNR \\
\midrule
        3DGS  & 2.14 & 1.53 & 2.08 & 1.68 & 3.49 & 2.21 & 1.43 & 2.07 & 2.22 & 1.75 & 1.79 & 2.55 & 1.53 & 1.52 & 1.50 & 1.96 & \cellcolor{myorange}32.82 \\
        SUGAR & 1.47 & 1.33 & 1.13 & 0.61 & 2.25 & 1.71 & 1.15 & 1.63 & 1.62 & 1.07 & 0.79 & 2.45 & 0.98 & 0.88 & 0.79 & 1.33 & 31.59 \\
        2DGS  & \cellcolor{myred}0.48 & \cellcolor{myyellow}0.91 & \cellcolor{myred}0.39 & \cellcolor{myred}0.39 & \cellcolor{myred}1.01 & \cellcolor{myyellow}0.83 & \cellcolor{myorange}0.81 & \cellcolor{myyellow}1.36 & \cellcolor{myyellow}1.27 & \cellcolor{myyellow}0.76 & \cellcolor{myred}0.70 & \cellcolor{myyellow}1.40 & \cellcolor{myred}0.40 & \cellcolor{myyellow}0.76 & \cellcolor{myred}0.52 & \cellcolor{myorange}0.80 & 32.43 \\
        GOF   & \cellcolor{myorange}0.50 & \cellcolor{myred}0.82 & \cellcolor{myorange}0.37 & \cellcolor{myorange}0.37 & \cellcolor{myyellow}1.12 & \cellcolor{myred}0.78 & \cellcolor{myred}0.73 & \cellcolor{myred}1.18 & \cellcolor{myred}1.29 & \cellcolor{myred}0.71 & \cellcolor{myorange}0.77 & \cellcolor{myred}0.90 & \cellcolor{myorange}0.44 & \cellcolor{myred}0.69 & \cellcolor{myorange}0.49 & \cellcolor{myred}0.74 & \cellcolor{myyellow}32.58 \\
        Ours & \cellcolor{myyellow}0.65 & \cellcolor{myorange}0.77 & \cellcolor{myyellow}0.58 & \cellcolor{myyellow}0.53 & \cellcolor{myorange}1.08 & \cellcolor{myorange}1.01 & \cellcolor{myyellow}0.96 & \cellcolor{myorange}1.31 & \cellcolor{myorange}1.45 & \cellcolor{myorange}0.72 & \cellcolor{myyellow}0.88 & \cellcolor{myorange}1.53 & \cellcolor{myyellow}0.67 & \cellcolor{myorange}0.83 & \cellcolor{myyellow}0.66 & \cellcolor{myyellow}0.91 & \cellcolor{myred}33.65 \\
        \bottomrule
    \end{tabular}
    }
\label{tab:dtucd}
\end{table*}

\noindent\textbf{Efficiency.}
Table~\ref{tab:eff} compares the model size and training time of \method with 3DGS, 2DGS, and GaussianPro on LLFF and Tanks\&Temples. While 2DGS uses the fewest Gaussians, its training time exceeds that of 3DGS. GaussianPro enhances appearance quality over 3DGS but incurs significantly higher training cost. In contrast, \method strikes a favorable balance, requiring fewer Gaussians than both 3DGS and GaussianPro, while achieving the shortest training time among all methods.

\begin{figure}[h]
    \centering
    \begin{minipage}{0.4\textwidth}
        \centering
        \scriptsize
        \captionsetup{skip=1pt}
        \setlength{\tabcolsep}{4pt}
        \captionof{table}{\small Comparison on efficiency. \# Gaussians is the average number of Gaussians.}
        \begin{tabular}{ccccc}
        \toprule
        Dataset & Method & PSNR & \#Gaussians & Training \\
        \midrule
        \multirow{4}{*}{LLFF}
        & 3DGS & 26.12 & 919K & 10min\\
        & 2DGS & 24.93 & \cellcolor{myred}343K & 11min\\
        & GaussainPro & \cellcolor{myorange}26.53 & 933K & 20min\\
        & \method & \cellcolor{myred}27.34 & \cellcolor{myorange}581K & \cellcolor{myred}9min\\
        \midrule
        & 3DGS & 23.85 & 1502K & 13min\\
        Tanks\& & 2DGS & 22.96 & \cellcolor{myred}416K & 15min\\
        Temples & GaussainPro  & \cellcolor{myorange}23.92 & 1381K & 35min\\
         & \method & \cellcolor{myred}24.41 & \cellcolor{myorange}754K & \cellcolor{myred}11min\\
        \bottomrule
        \end{tabular}
        \label{tab:eff}
    \end{minipage}\hfill
    \begin{minipage}{0.5\textwidth}
        \centering
        \scriptsize
        \captionsetup{skip=2pt}
        \setlength{\tabcolsep}{5pt}
        \captionof{table}{\small Ablation study. Repr. stands for the Gaussian type, Hyb. for hybrid rasterization, Ex. for type exchange, and Freq. for frequency regularization.}
        \begin{tabular}{c c c c c|ccc}
        \toprule
        ID & Repr. & Hyb.  & Ex. & Freq. & PSNR$\uparrow$ & SSIM $\uparrow$ & LPIPS $\downarrow$ \\
        \midrule
        \textit{i} & 3D&\xmark & \xmark & \xmark & 26.12 & 0.865 & 0.099\\
        \textit{ii} & 3D+2D& \xmark & \xmark & \xmark & 26.01 & 0.859 & 0.105\\
        \textit{iii} & 3D+2D& \cmark  & \xmark & \xmark & 26.23 & 0.867 & 0.097\\
        \textit{iv} & 3D+2D& \cmark  & \cmark & \xmark & \cellcolor{myorange}26.58 & \cellcolor{myorange}0.874 & \cellcolor{myorange}0.093\\
        \textit{v}& 3D & \xmark  & \xmark & \cmark &26.19 &0.867 &0.101 \\
        \textit{vi}& 3D+2D & \cmark  & \xmark & \cmark & \cellcolor{myyellow}26.41 & \cellcolor{myyellow}0.871 & \cellcolor{myyellow}0.096\\
        \textit{vii}& 3D+2D & \cmark & \cmark  & \cmark & \cellcolor{myred}27.34 & \cellcolor{myred}0.895 & \cellcolor{myred}0.083\\
        \bottomrule
        \end{tabular}
        \label{tab:ablate}
    \end{minipage}\hfill
    \vspace{-0.18in}
\end{figure}

\subsection{Ablation and Generalization Analysis}\label{sec:ablation}
\noindent\textbf{Ablation Study.} 
We evaluate the effectiveness of each component in \method in Table~\ref{tab:ablate}. Row~\textit{i} is the vanilla 3DGS baseline. In row~\textit{ii}, we adopt a hybrid 2D/3D representation but rasterize all Gaussians using the 3DGS rasterizer, which leads to performance degradation. Row~\textit{iii} incorporates our hybrid rasterizer, which renders Gaussians according to their type. However, this setting still lacks flexibility and regularization. Row~\textit{iv} incorporates adaptive type exchange to enhance the flexibility. Rows~\textit{v}–\textit{vii} study frequency-based supervision, which provides only limited gains for non-hybrid 3DGS (row~\textit{v}) but is more effective in the hybrid setting. The full model in row~\textit{vii} achieves the best performance, indicating that decoupled frequencies more effectively exploit the strengths of the exchangeable hybrid representation. We provide more ablation and analysis in Appendix~\ref{app:proj}.


\begin{wrapfigure}{r}{0.4\textwidth}
    \vspace{-0.2in}
    \centering
    \captionsetup{skip=2pt}
    \begin{minipage}{\linewidth}
        \centering
        \includegraphics[width=0.9\linewidth]{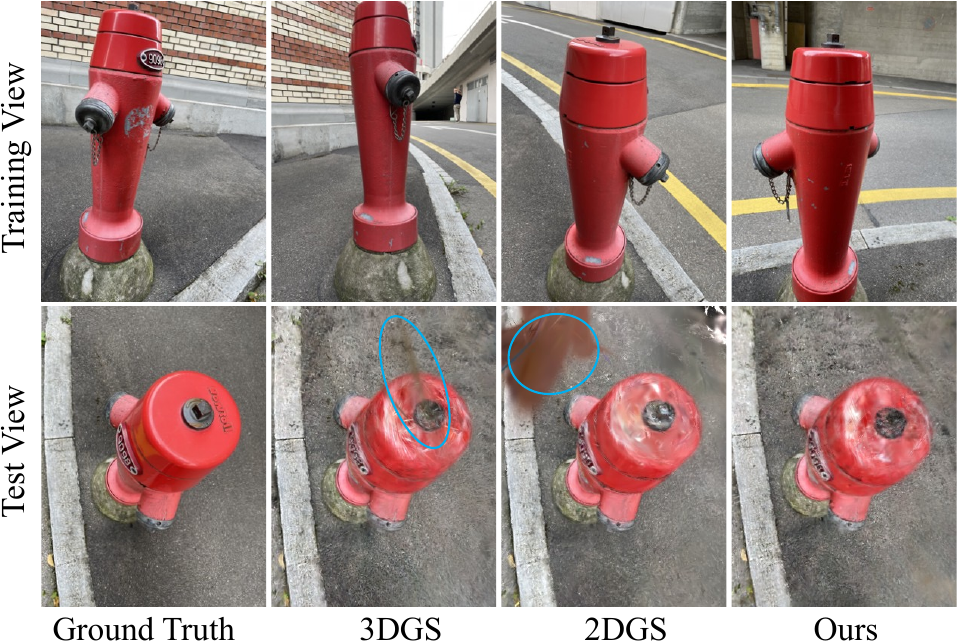}
        \caption{\small Comparison in the OOD setting.}
        \label{fig:ood}
    \end{minipage}
    \begin{minipage}{\linewidth}
        \footnotesize
        \centering
        \setlength{\tabcolsep}{2pt}
        \captionof{table}{ \small Generalization performance.}
        \label{tab:ood}
        \resizebox{0.95\linewidth}{!}{
        \begin{tabular}{ccccc}
        \toprule
        Setting & Method & PSNR $\uparrow$ & SSIM $\uparrow$ & LPIPS $\downarrow$ \\
        \midrule
        \multirow{3}{*}{Few-shot}
        & 3DGS & 19.52 & 0.719 & 0.279 \\
        & 2DGS & 18.50 & 0.661 & 0.321 \\
        & \method &\textbf{20.13} &\textbf{0.735} &\textbf{0.258}  \\
        \midrule
        \multirow{3}{*}{OOD}
        & 3DGS & 20.21 &0.763 &0.242 \\
        & 2DGS &23.52  &0.863 &0.188 \\
        & \method &\textbf{24.17} &\textbf{0.907} &\textbf{0.151} \\
        \bottomrule
        \label{tab:ood}
        \end{tabular}
        }
    \end{minipage}
    \vspace{-0.5in}
\end{wrapfigure}
\noindent\textbf{Generalization Analysis.} We evaluate the robustness of \method in challenging scenarios, including few-shot and out-of-distribution (OOD) settings. Following prior work, we use LLFF~\cite{mildenhall2019llff} for few-shot evaluation~\cite{li2024dngaussian, zhu2024fsgs} and OOD-NVS~\cite{chen2025splatformer} for OOD evaluation~\cite{yu2024mipsplat}. More details are provided in Appendix~\ref{app:details}. As shown in Table~\ref{tab:ood} and Figure~\ref{fig:ood}, \method achieves robust performance in both settings, benefiting from its balanced multi-view consistency and appearance fidelity. This indicates that the hybrid representation generalizes better than single-type baselines.

We also emphasize that \method serves as a general underlying representation and is compatible with various optimization strategies developed for specialized settings~\cite{li2024dngaussian, zhu2024fsgs, yin2024fewviewgs, xu2024mvpgs, shen2024solidgs, chen2025splatformer}. We discuss these orthogonal techniques in Appendix~\ref{app:related}, and provide further remarks on limitations and broader impacts in Appendix~\ref{app:diss}.


%% file: sec/5_conclusion.tex
\section{Conclusion}
\label{sec:conc}


This paper presents \method, a hybrid Gaussian Splatting framework that combines the appearance fidelity of 3D Gaussians with the geometric accuracy of 2D Gaussians. The design integrates Hybrid Gaussian Rasterization for unified rendering, Adaptive Type Exchange for flexible representation, and Frequency-Decoupled Optimization to balance geometry and appearance. \method outperforms both 2D- and 3D-only baselines across multiple benchmarks. Future work includes extending the hybrid representation to more diverse and challenging scenarios.

\textbf{Acknowledgments}: This work was supported by Intelligence Advanced Research Projects Activity (IARPA) via Department of Interior/Interior Business
Center (DOI/IBC) contract number 140D0423C0074. The
U.S. Government is authorized to reproduce and distribute
reprints for Governmental purposes, notwithstanding any
copyright annotation thereon. Disclaimer: The views
and conclusions contained herein are those of the authors
and should not be interpreted as necessarily representing
the official policies or endorsements, either expressed or
implied, of IARPA, DOI/IBC, or the U.S. Government.

%% file: sec/6_appendix.tex
\addtocounter{equation}{2}
Table of contents:
\begin{itemize}
    \item Appendix~\ref{app:details}: \textbf{Implementation Details.} Additional information on the training pipeline, parameter settings, and evaluation datasets and protocols used in our experiments.

    \item Appendix~\ref{app:related}: \textbf{Related Works.} Extended discussion on the distinctions between our method and other related approaches, including hybrid and task-specific splatting methods.

    \item Appendix~\ref{app:rasterization}: \textbf{Differentiable Rasterization for 2D/3D Gaussian Splatting.} Technical details of the rasterization procedures used in 3DGS and 2DGS, which serve as the building blocks for our Hybrid Gaussian Rasterization module.

    \item Appendix~\ref{app:repa}: \textbf{Effective Rank and Adaptive Type Exchange.} More ablations and analysis on the effective rank threshold. We also detail the design of permutation-based reparameterization, and report the evolution and distribution of Gaussian types during training.

    \item Appendix~\ref{app:dwt}: \textbf{Discrete Wavelet Transform.} A detailed explanation of the DWT used in our Frequency-Decoupled Optimization, including mathematical formulation and visualization of decomposed frequency components.

    \item Appendix~\ref{app:proj}: \textbf{Gradient Conflict Analysis in Frequency-Decoupled Optimization.} In-depth analysis of gradient conflicts arising from different frequency components, supported by empirical statistics. We also compare different loss application strategies and highlight the benefits of our projection-based solution.

    \item Appendix~\ref{app:diss}: \textbf{Discussion.} Discussion on limitations, broader impacts, and the generalization potential of our hybrid representation in more diverse scenarios.
\end{itemize}

\section{Implementation Details}
\label{app:details}

\noindent\textbf{Training Pipeline and Parameter Setting.} 
Our training setup closely follows 3DGS~\cite{3dgs}. We assume camera poses are provided or can be estimated using structure-from-motion (SfM)~\cite{schonberger2016colmap}. Initial sparse point clouds are generated via COLMAP~\cite{schonberger2016colmap, schonberger2016colmap2}. All methods, including \method and baselines, are trained for 30K iterations. Learning rates for Gaussian parameters follow the default configurations from 3DGS and 2DGS. We adopt the densification strategy from 3DGS, which refines Gaussian distributions by pruning or duplicating them in under- or over-reconstructed regions. Densification begins at iteration 500, ends at iteration 15K, and is performed every 100 iterations.

In \method, we employ Hybrid Gaussian Rasterization, where each Gaussian is rendered according to its assigned type. Gaussian types are randomly initialized. To enable flexible representation, we introduce Adaptive Type Exchange, which is performed every 500 iterations from step 500 to 30K. During type switching, we set the effective rank threshold $\theta_e$ to 2.05. For scale modulation, which allows 2D Gaussians to evolve into 3D Gaussians as the $s_z$ scale becomes more significant, we use $\theta_z = 1.05$ and temperature $T = 0.001$ in Eq.~\ref{e:soft}. Additionally, we incorporate frequency-based supervision using the Discrete Wavelet Transform (DWT). More details on Frequency-Decoupled Optimization are provided in Appendix~\ref{app:proj}.

\textbf{Datasets and Evaluation Protocols.}
We evaluate the performance of \method and baselines on several widely used datasets, including Mip-NeRF360~\cite{barron2022mip}, LLFF~\cite{mildenhall2019llff}, Tanks\&Temples~\cite{knapitsch2017tanks}, DTU~\cite{jensen2014dtu}, and OOD-NVS~\cite{chen2025splatformer}. We follow the standard train/test splits used in prior work~\cite{barron2022mip}, with additional dataset statistics provided in Table~\ref{tab:datasets}. To ensure fair comparison with baselines~\cite{3dgs, 2dgs, yu2024gof}, we downsample input images using the same factors as in~\cite{yu2024mipsplat}. All datasets provide RGB images for evaluating appearance quality. DTU additionally offers ground-truth point clouds for computing Chamfer Distance, while Tanks\&Temples provides ground-truth depth maps for depth accuracy evaluation. Beyond standard dense-view evaluation, we also perform evaluation under few-shot settings using 3 views from LLFF, and out-of-distribution (OOD) setting using OOD-NVS.

\begin{table}[h]
\centering
\scriptsize
\caption{Details on the datasets used for evaluation of appearance and geometry.}
    \begin{tabular}{l c c c c}
    \toprule
    \textbf{Dataset} & \textbf{Ground Truth} & \textbf{Evaluation Metric} &\textbf{Evaluation Protocol} & \textbf{Factor} \\
    \midrule
    Mip-NeRF 360 (outdoor)~\cite{barron2022mip} & RGB image & Appearance &Standard & 4 \\
    Mip-NeRF 360 (indoor)~\cite{barron2022mip}  & RGB image & Appearance &Standard & 2 \\
    LLFF~\cite{mildenhall2019llff}                   & RGB image & Appearance &Standard and Few-shot & 8 \\
    Tanks\&Temples~\cite{knapitsch2017tanks}      & RGB image; Depth & Appearance and Geometry (F1) &Standard & 2 \\
    DTU~\cite{jensen2014dtu}                    & RGB image; Point Cloud & Geometry (chamfer distance) &Standard &  2 \\
    OOD-NVS~\cite{chen2025splatformer}              & RGB image & Appearance &OOD & 1 \\
    \bottomrule
    \end{tabular}
\label{tab:datasets}
\end{table}

\section{Related Works}
\label{app:related}
Following 3DGS~\cite{3dgs}, a line of work has been proposed to enhance the geometry accuracy and reconstruction quality. As discussed in the related work section, most existing methods are based on a single representation~\cite{guedon2024sugar, cheng2024gaussianpro, 2dgs, dai2024surfles, yu2024gof, xu2024texturegs, zhang2024fregs}. Here, we further expand on several recent efforts that aim to exploit the advantages of 3D and 2D Gaussians, as summarized in Table~\ref{tab:hygs}. We consider a method to be general if the training pipeline follows the standard 3DGS.

HybridGS~\cite{lin2024hybridgs} proposes to combine 3D Gaussians and image-space 2D Gaussians to remove transient objects during reconstruction. However, the 2D Gaussians in HybridGS are defined in the image frame, lacking the multi-view consistency provided by 2DGS~\cite{2dgs}. Additionally, HybridGS employs a three-stage training pipeline specifically designed for transient object removal, making it less general than pipelines based on 3DGS. Moreover, since the training code of HybridGS is not publicly available, a direct comparison with \method is not feasible.

HorizonGS~\cite{jiang2024horizon}, on the other hand, is built upon ScaffoldGS, where voxel-based MLPs are used to decode Gaussian primitives. HorizonGS generates 3D Gaussians from MLPs for novel view synthesis and 2D Gaussians for surface reconstruction. Thus, it still follows a single-representation scheme. Furthermore, HorizonGS is specifically designed for aerial-ground scenarios and introduces a two-stage training pipeline to address conflicts from varying altitudes. Similar to HybridGS, this design is task-specific and less generalizable.

In contrast to HybridGS and HorizonGS, we note that several recent optimization techniques are more general and adhere to the 3DGS training pipeline, such as ScaffoldGS~\cite{lu2024scaffold} and 3DGS-MCMC~\cite{kheradmand2024mcmc}. ScaffoldGS improves efficiency by introducing voxel-based MLPs, while 3DGS-MCMC enhances the densification process by reformulating 3D Gaussians as Markov Chain Monte Carlo (MCMC) samples. These methods are orthogonal to \method. Since \method serves as a general underlying representation, we believe such orthogonal optimizations can be incorporated to further enhance our framework. Exploring the potential of integrating these techniques with our exchangeable hybrid representation is a promising direction for future work.

In this work, we assume that camera poses are either available or can be estimated using structure-from-motion (SfM)~\cite{schonberger2016colmap}, and that initial sparse point clouds can be generated using COLMAP~\cite{schonberger2016colmap, schonberger2016colmap2}. However, in practice, recovering geometric information such as camera poses and point clouds remains challenging. Recently, 3D Geometric Foundation Models (GFMs) have emerged as a promising approach to improve the generalizability of 3D reconstruction~\cite{cong2025e3d}. Feed-forward models such as DUSt3R~\cite{wang2024dust3r}, MASt3R~\cite{leroy2024mast3r}, and VGGT~\cite{wang2025vggt} can predict robust geometric attributes in a single forward pass, even when the input multi-view images exhibit minimal or no overlap. Incorporating GFMs can potentially enhance the reconstruction quality of our method in open-world scenarios.

\begin{table}[h]
\centering
\footnotesize
\caption{Comparison of the setting with related works.}
    \begin{tabular}{l c c c}
    \toprule
    \textbf{Method} & \textbf{Gaussian Type} & \textbf{Setting} & \textbf{Training Pineline} \\
    \midrule
    3DGS~\cite{3dgs}  &3D    & General &Single-stage \\
    2DGS~\cite{2dgs}  &2D    & General &Single-stage \\
    HybridGS~\cite{lin2024hybridgs} &3D+2D & Transient &Three-stage \\
    HorizonGS~\cite{jiang2024horizon}  &3D/2D & Varying-altitude &Two-stage  \\
    ScaffoldGS~\cite{lu2024scaffold}   &3D   & General &Single-stage \\
    3DGS-MCMC~\cite{kheradmand2024mcmc} &3D    & General &Single-stage \\
    \midrule
    \method &3D+2D  & General &Single-stage\\
    \bottomrule
    \end{tabular}
\label{tab:hygs}
\end{table}

\section{Differentiable Rasterization for 2D/3D Gaussians Splatting }
\label{app:rasterization}
In this section, we detail the differentiable rasterization procedures used in 3DGS~\cite{3dgs} and 2DGS~\cite{2dgs}, focusing on how the intermediate coordinates in Eq.(\ref{e:dist2}) are computed. In both methods, images are rendered by computing the color at each screen-space pixel $x_p$ from a set of $N$ Gaussians $\{\mathcal{G}_i\}_{i=0}^{N-1}$. The final pixel color is obtained by rasterizing the Gaussians onto the image plane, but the rasterization procedures differ significantly between 3DGS and 2DGS.

3DGS employs an affine approximation to the projective transformation for rasterization. For a 3D Gaussian $\mathcal{G}^{3d}_i$ with type $t_i = 1$, it is first projected onto the image plane via affine projection~\cite{3dgs}. The resulting projected Gaussian is denoted as $\mathcal{G}^{\text{proj}}_i$, with 2D center $\boldsymbol{\mu}'_{3d,i}$ and covariance $\boldsymbol{\Sigma}'_i$ in the image plane. The projected covariance is computed as:
\begin{equation}
\small
\boldsymbol{\Sigma}'_i = \boldsymbol{J} \boldsymbol{W} \boldsymbol{\Sigma}_i \boldsymbol{W}^T \boldsymbol{J}^T,
\end{equation}
where $\boldsymbol{J}$ is the Jacobian matrix of the affine projection, and $\boldsymbol{W}$ accounts for the world-to-camera transformation~\cite{3dgs}. Once $\boldsymbol{\Sigma}'_i$ is obtained, the projected center $\boldsymbol{\mu}'_{3d,i}$ can be computed accordingly.

On the other hand, 2DGS applies ray–splat–intersection to rasterize 2D Gaussians. For a 2D Gaussian $\mathcal{G}^{2d}_i$ with $t_i = 0$, it is not directly projected onto the image plane. Instead, to preserve the geometric accuracy of $\mathcal{G}^{2d}_i$, the pixel $x_p$ is unprojected into the local tangent frame defined by the Gaussian~\cite{2dgs}. This is done by computing the intersection between the ray passing through $x_p$ and the tangent plane of $\mathcal{G}^{2d}_i$, resulting in the local coordinates:
\begin{equation}
\label{e:uv}
\small
u(x_p) = \frac{h_u^2 h_v^4 - h_u^4 h_v^2}{h_u^1 h_v^2 - h_u^2 h_v^1}, \quad 
v(x_p) = \frac{h_u^4 h_v^1 - h_u^1 h_v^4}{h_u^1 h_v^2 - h_u^2 h_v^1},
\end{equation}
where $h_u$ and $h_v$ are derived from the homogeneous plane equations associated with the pixel $x_p = (x, y)$ as:
\begin{equation}
\small
h_u = (\boldsymbol{WH})^T h_x, \quad h_v = (\boldsymbol{WH})^T h_y.
\end{equation}

More details about the homogeneous transformation matrices $\boldsymbol{H}$ and $\boldsymbol{W}$ can be found in 2DGS~\cite{2dgs}. By solving Eq.~(\ref{e:uv}), we obtain the 2D position of $x_p$ in the tangent frame, denoted as $\mathbf{u}_i(x_p) = (\boldsymbol{u}_i(x_p), \boldsymbol{v}_i(x_p))$. Note that the center $\boldsymbol{\mu}_{2d,i}$ of $\mathcal{G}_i^{2d}$ is defined as the origin of this tangent frame. 

With the projected center $\boldsymbol{\mu}'_{3d,i}$ and covariance $\boldsymbol{\Sigma}_i$ for 3D Gaussians, and the intersection coordinates $\boldsymbol{u}_i(x_p)$ and $\boldsymbol{v}_i(x_p)$ for 2D Gaussians, we can perform hybrid rasterization as described in Section~\ref{sec:hyras}.

\section{Effective Rank and Type Switch}
\label{app:repa}
In this section, we provide additional details about Adaptive Type Exchange. We focus on the effective
\begin{wraptable}{r}{0.5\textwidth}
    \centering
    \vspace{-0.1in}
    \footnotesize
    \caption{Ablation on different erank thresholds. We evaluate the appearance of \method on the LLFF dataset with different erank thresholds.}
    \vspace{-2mm}
    \begin{tabular}{c c c c}
    \toprule
    \textbf{erank threshold} & \textbf{PSNR} & \textbf{SSIM} & \textbf{LPIPS} \\
    \midrule
    1.9  & 26.15 & 0.871 & 0.103 \\
    1.95 & 26.23 & 0.868 & 0.101 \\
    2    & 26.49 & 0.874 & 0.097 \\
    2.05 & 27.24 & 0.885 & 0.086 \\
    2.1  & 26.37 & 0.844 & 0.113 \\
    2.15 & 25.78 & 0.831 & 0.126 \\
    \bottomrule
    \end{tabular}
    \label{tab:erank}
    \vspace{-0.2in}
\end{wraptable}
rank threshold and the design choices behind 3D Gaussian reparameterization and 2D Gaussian scale modulation. We also present statistics on the distribution of Gaussian types over training iterations.

\textbf{Effective Rank Threshold.} As described in Section~\ref{sec:ats}, we use the effective rank (erank) to assess the mismatch between a Gaussian’s assigned type and its actual geometric dimensionality. We set the erank threshold to 2.05 in our experiments and study its effect in Table~\ref{tab:erank}. When a 3D Gaussian becomes increasingly flat and its erank drops below the threshold, it is converted to a 2D Gaussian. Conversely, if a 2D Gaussian’s erank exceeds the threshold, it is switched to 3D. A higher threshold causes more 3D Gaussians to be converted to 2D, as more will fall below the threshold. This can degrade performance when the threshold is set too high. In contrast, a lower threshold (e.g., 1.9) results in fewer 3D Gaussians being converted to 2D, causing the model to behave more like the 3DGS baseline and limiting the benefits of hybrid representation.

While the erank metric has been previously introduced~\cite{roy2007erank, hyung2024erankgs}, our contribution lies in its integration into a dynamic type exchange mechanism for hybrid Gaussian representation. We acknowledge that erank is a heuristic measure of effective dimensionality and may not behave monotonically in all scenarios. When a Gaussian has scales $(1,1,s_z)$ with the first two scales fixed, the erank increases with $s_z$ initially but may drop as $s_z$ becomes dominant (e.g., erank returns to 2 when $s_z \approx 2.6$). In such cases, a volumetric Gaussian could technically fall below the threshold $\theta_e$ and be converted to 2D. However, we note that such configurations are rare in practice. All three scales are updated jointly during training, and our method includes reparameterization and soft modulation to ensure that type transitions remain stable and consistent with the evolving shape. Although the erank threshold does not come with theoretical guarantees for all edge cases, it demonstrates empirical effectiveness across diverse datasets.

\begin{figure*}[h]
    \centering
    \includegraphics[width=1\linewidth]{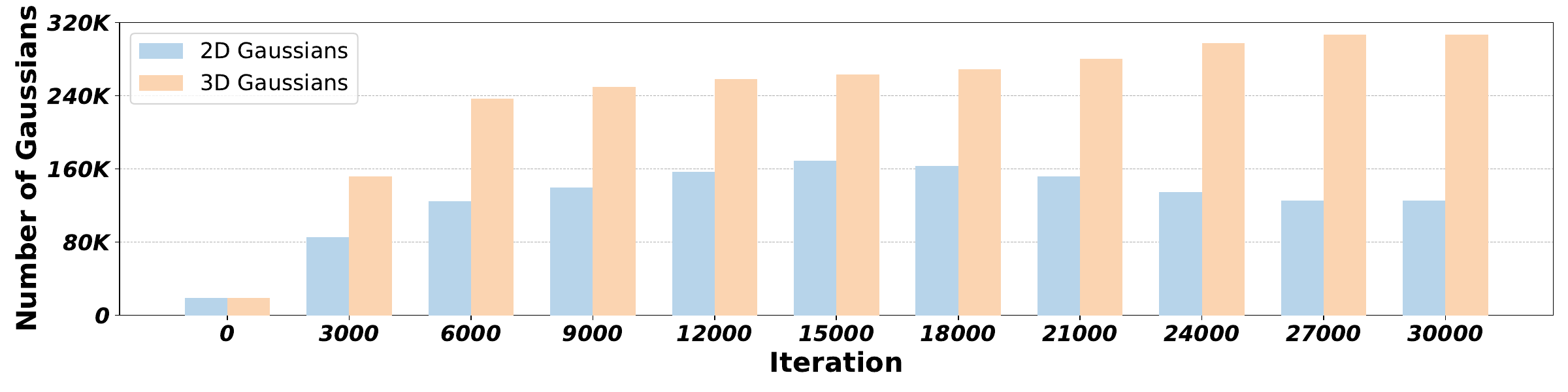}
    \caption{\small The number 3D and 2D Gaussians in different iterations during training.}
    \label{fig:cnt}
\end{figure*}

\textbf{Permutation-Based Reparameterization.}
The key to reparameterization is preserving the covariance during type switching. We achieve this using permutation matrices $\boldsymbol{P}_x$ and $\boldsymbol{P}_y$, designed to be orthogonal with a positive determinant. Specifically, we define:
\begin{equation}
\small
\boldsymbol{P}_x = \begin{bmatrix}
    0 & 1 & 0 \\
    0 & 0 & 1 \\
    1 & 0 & 0
\end{bmatrix}, \quad
\boldsymbol{P}_y = \begin{bmatrix}
    0 & 0 & 1 \\
    1 & 0 & 0 \\
    0 & 1 & 0
\end{bmatrix}
\end{equation}
where $\boldsymbol{P}\boldsymbol{P}^T = \boldsymbol{I}$ and $\boldsymbol{I}$ is the identity matrix, and $\text{det}(\boldsymbol{P})=1$. During conversion, the new $\boldsymbol{S}^*$ scaling and rotation $\boldsymbol{R}^*$ are computed using these permutation matrices to ensure that the transformed covariance $\boldsymbol{\Sigma} = \boldsymbol{RSSR}^T$ remains unchanged:
\begin{equation}
     \boldsymbol{S}^* = \boldsymbol{P}\boldsymbol{S}\boldsymbol{P} ^T\text{ and } \boldsymbol{R}^* = \boldsymbol{R}\boldsymbol{P}^T
\end{equation}
It is easy to see the converted covariance is unchanged:
\begin{align}
\boldsymbol{\Sigma}^* 
    &= \boldsymbol{R}^* \boldsymbol{S}^{*} \boldsymbol{S}^{*T} \boldsymbol{R}^{*T} \notag \\
    &= \boldsymbol{R}\boldsymbol{P}^T \cdot \boldsymbol{P}\boldsymbol{S}\boldsymbol{P}^T \cdot \boldsymbol{P}\boldsymbol{S^T}\boldsymbol{P^T} \cdot \boldsymbol{P}\boldsymbol{R}^T \notag \\
    &= \boldsymbol{R} \cdot (\boldsymbol{P}^T \boldsymbol{P}) \cdot \boldsymbol{S} \cdot (\boldsymbol{P}^T \boldsymbol{P}) \cdot \boldsymbol{S}^T \cdot (\boldsymbol{P}^T \boldsymbol{P}) \cdot \boldsymbol{R}^T \notag \\
    & = \boldsymbol{RSS^TR^T} \notag\\
    &= \boldsymbol{\Sigma}
\label{eq:sigma_star}
\end{align}

In addition to preserving the covariance, the permutation must ensure that the converted rotation matrix $\boldsymbol{R}^*$ has a positive determinant. This is important because $\boldsymbol{R}^*$ is converted into a quaternion during optimization, and most 3D graphics frameworks assume right-handed coordinate systems~\cite{barrow1978cd}. A negative determinant implies a reflection, which cannot be represented by a unit quaternion. Given that the original rotation matrix $\boldsymbol{R}$ satisfies $\det(\boldsymbol{R}) > 0$, and the permutation matrices $\boldsymbol{P}_x$ and $\boldsymbol{P}_y$ are orthogonal with $\det(\boldsymbol{P}) = 1$, the determinant of the converted rotation $\boldsymbol{R}^* = \boldsymbol{R} \boldsymbol{P}^T$ is given by:
$
\det(\boldsymbol{R}^*) = \det(\boldsymbol{R}) \cdot \det(\boldsymbol{P}^T) = \det(\boldsymbol{R}) \cdot \det(\boldsymbol{P}) = \det(\boldsymbol{R}) > 0.
$ Therefore, $\boldsymbol{P}_x$ and $\boldsymbol{P}_y$ preserve both the covariance and the positive determinant of the rotation matrix, ensuring compatibility with quaternion-based optimization.

\textbf{Gaussian-Type Distribution.}
Figure~\ref{fig:cnt} shows the distribution of 2D and 3D Gaussians throughout training. As described in Appendix~\ref{app:details}, we randomly initialize Gaussian types, resulting in roughly equal numbers of 2D and 3D Gaussians at the start. During optimization, both types increase due to densification, with a more significant rise in 3D Gaussians. This trend can be attributed to the SfM-initialized points already containing geometric structure, allowing 2D Gaussians to capture coarse geometry in early iterations. Notably, the number of 2D Gaussians gradually decreases in the later stages of training, especially after densification ends at 15K iterations. This suggests that many 2D Gaussians are being converted to 3D Gaussians, likely to better recover under-reconstructed regions and capture finer scene details.

\begin{table*}[b]
    \vspace{-0.25in}
    \centering
    \captionsetup{skip=1pt}
    \caption{\small Effect of 3D Gaussian percentage on reconstruction quality.
PSNR denotes the peak signal-to-noise ratio at each iteration step under different initialization strategies.}
    \label{tab:init}
    \resizebox{1\linewidth}{!}{
    \begin{tabular}{lcccccccccccc}
        \toprule
        Iter. & 0 & 3000 & 6000 & 9000 & 12000 & 15000 & 18000 & 21000 & 24000 & 27000 & 30000 & PSNR \\
        \midrule
        All 2D initialization & 0.0\% & 19.0\% & 29.1\% & 33.7\% & 35.0\% & 38.9\% & 39.8\% & 41.4\% & 43.2\% & 45.7\% & 47.8\% & 27.25 \\
        All 3D initialization & 100.0\% & 86.2\% & 75.3\% & 69.0\% & 63.9\% & 59.7\% & 58.7\% & 58.0\% & 57.4\% & 57.0\% & 57.0\% & 27.51 \\
        Random initialization & 49.9\% & 58.1\% & 59.9\% & 57.3\% & 55.1\% & 52.3\% & 52.5\% & 52.9\% & 52.9\% & 53.1\% & 54.2\% & 27.86 \\
        \bottomrule
    \end{tabular}
    }
    \vspace{-0.12in}
\end{table*}

\textbf{Gaussian-Type Initialization.} A key feature of EGGS is its exchangeable representation, which allows each Gaussian primitive to change its type as needed, regardless of its initial type. To verify this capability, we conduct a simple experiment by investigating three initialization scenarios: we initialize all Gaussians as 2D, all as 3D, or with random type assignments, and observe the distribution of Gaussian types throughout training. Below, we show the percentage of 3D Gaussians at different iterations. As shown in Table~\ref{tab:init}, even when the model is initialized entirely with 2D Gaussians, part of the Gaussians are converted to 3D Gaussians during training, leading to a hybrid model in the final stage. This demonstrates that 2D Gaussians can indeed transition to 3D types during training, despite potentially incorrect initialization, and vice versa.

\section{Discrete Wavalet Transformation}
\label{app:dwt}
\begin{figure*}[h]
    \centering
    \includegraphics[width=0.8\linewidth]{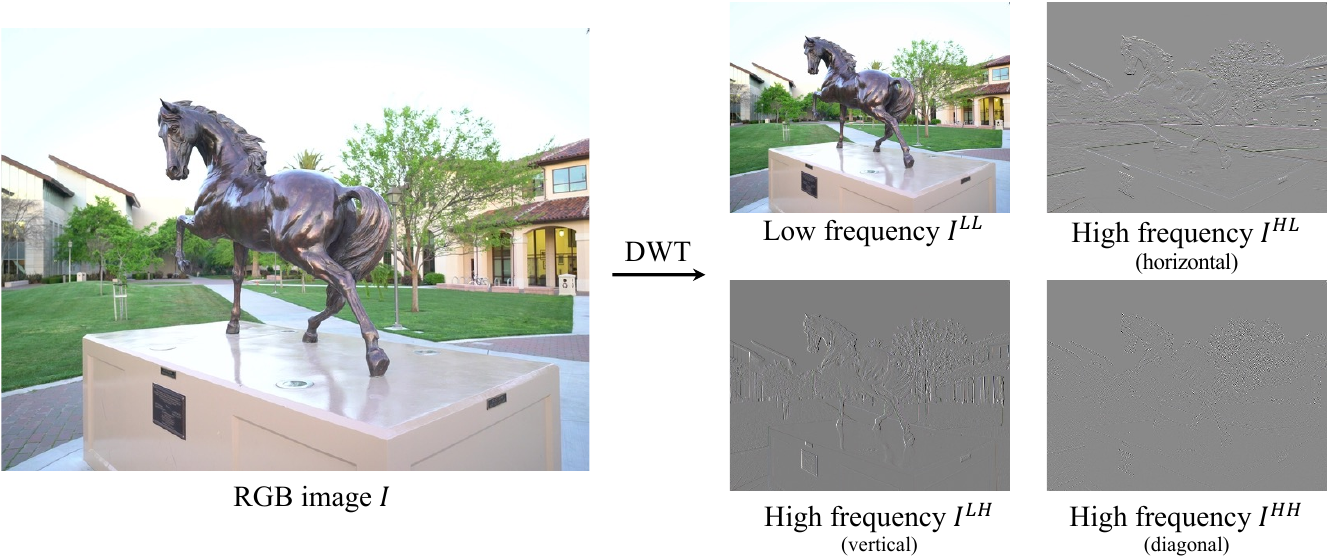}
    \caption{\small Illustration of the level-1 Discrete Wavalet Transformation.}
    \label{fig:dwtill}
\end{figure*}
In this section, we provide additional details on the Discrete Wavelet Transform (DWT) used in our Frequency-Decoupled Optimization. DWT is a widely adopted technique for frequency-domain analysis. Given an image $\mathcal{I}$, DWT decomposes it into four sub-bands: one low-frequency component and three high-frequency components corresponding to horizontal, vertical, and diagonal directions. Formally, we define the low-pass filter matrix $\boldsymbol{L}$ as:
\begin{equation}
\textstyle
\small
    L = \begin{bmatrix}
    \cdots & \cdots & \cdots & & \\
    \cdots & \ell_{-1} & \ell_0 & \ell_{1} & \cdots & \\
    \cdots & \cdots & \ell_{-1} & \ell_0 & \ell_{1} & \cdots \\
    &&& \cdots & \cdots & \cdots \\
    \end{bmatrix}
\end{equation}
where $\boldsymbol{\ell}$ is the 1D low-pass wavelet filter. Similarly, the high-pass matrix $\boldsymbol{H}$ is derived from the 1D high-pass wavelet filter $\boldsymbol{h}$. We use orthogonal 1D wavelet filters such that $\boldsymbol{\ell}$ and $\boldsymbol{h}$ are the same~\cite{heil1989dwt, strang1996wavelets}. The four sub-bands are computed as:
\begin{align}
\mathcal{I}^{LL} &= L\mathcal{I}L^T; \notag\\
\mathcal{I}^{LH} &= H\mathcal{I}L^T; \notag\\
\mathcal{I}^{HL} &= L\mathcal{I}H^T; \notag\\
\mathcal{I}^{HH} &= H\mathcal{I}H^T;
\end{align}
We provide illustrative example in Figure~\ref{fig:dwtill}. We extract the low-frequency feature as $\mathcal{I}_l = \mathcal{I}^{LL}$, and the high-frequency component $\mathcal{I}_h$ as the composition of directional details: $\mathcal{I}^{LH}$ (horizontal), $\mathcal{I}^{HL}$ (vertical), and $\mathcal{I}^{HH}$ (diagonal). In our implementation, we use a level-1 Haar filter for the DWT.

\section{Gradient Conflict Analysis in Frequency-Decoupled Optimization.}

\label{app:proj}

We provide additional details on Frequency-Decoupled Optimization and Algorithm~\ref{alg:fdo}. After applying DWT, both 3D and 2D Gaussians receive gradients from the high-frequency and low-frequency losses. As discussed in Section~\ref{sec:wave}, the distinct roles of 3D and 2D Gaussians—where 3D Gaussians prioritize fine detail and 2D Gaussians emphasize geometric structure—can lead to conflicting gradient directions.

\begin{figure*}[t]
    \centering
    \includegraphics[width=1\linewidth]{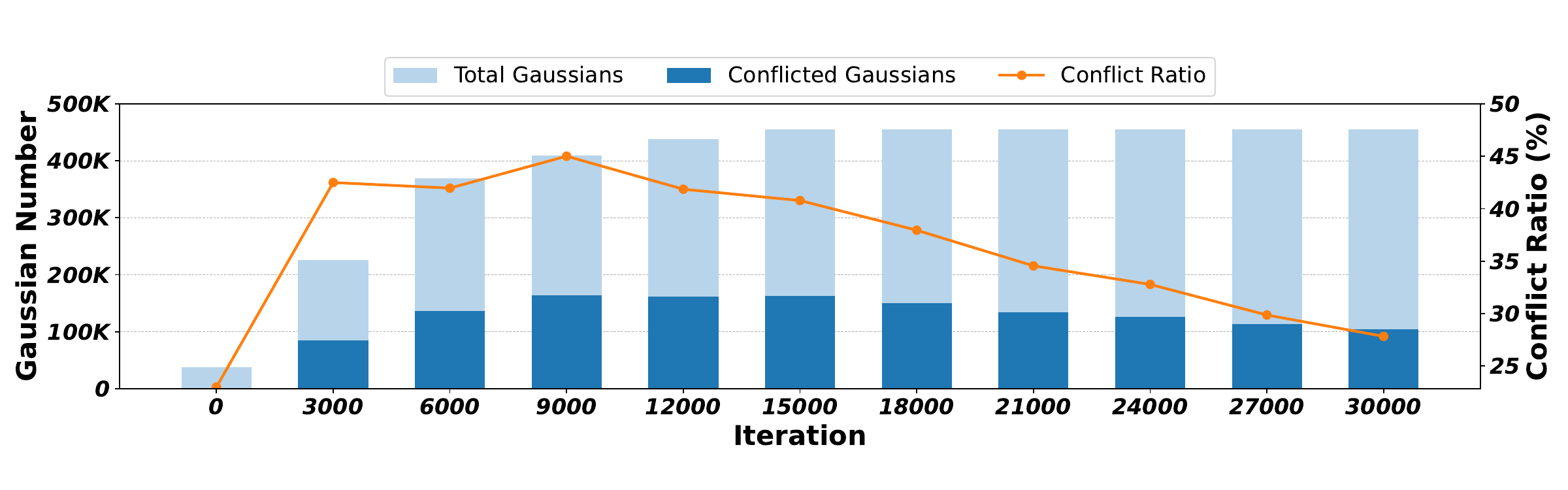}
    \caption{\small The number of Gaussians with conflicted gradients in different iterations during training.}
    \label{fig:conf}
\end{figure*}

We empirically analyze this in Figure~\ref{fig:conf}, where “conflicted Gaussians” are defined as those with negative inner product between low- and high-frequency gradients, i.e., $\mathbf{g}^{\text{low}}_i \cdot \mathbf{g}^{\text{high}}_i < 0$ in Algorithm~\ref{alg:fdo}. In early training (e.g., before 15K iterations), about 45\% of Gaussians exhibit such conflicts.
\begin{wraptable}{r}{0.5\textwidth}
    \centering
    \footnotesize
    \caption{\small Ablation on different use of the frequency loss. FDO stands for Frequency-Decoupled Optimization. }
    \vspace{-2mm}
    \begin{tabular}{l c c c}
    \toprule
    \textbf{Method} & \textbf{PSNR} & \textbf{SSIM} & \textbf{LPIPS} \\
    \midrule
    3DGS  & 26.12 & 0.865 & 0.099 \\
    \method w/o FDO & 26.58 & 0.874 & 0.093 \\
    \method w/ DWT    & 26.82 & 0.877 & 0.092 \\
    \method w/ DWT + mask & 27.07 & 0.881 & 0.089 \\
    \method  & 27.34 & 0.895 & 0.083 \\
    \bottomrule
    \end{tabular}
    \label{tab:pc}
    \vspace{-0.2in}
\end{wraptable}
Although the conflict ratio gradually decreases as training progresses, around 20\% of Gaussians still experience conflicts at convergence. This supports our motivation that naively combining $\mathcal{L}_{\text{low}}$ and $\mathcal{L}_{\text{high}}$ as Eq.(\ref{e:loss}) for all Gaussians provides suboptimal supervision, and underscores the need for the proposed asymmetrical update strategy.
In Table~\ref{tab:pc}, we compare different strategies for applying frequency-based supervision. As baselines, we include vanilla 3DGS and \method without Frequency-Decoupled Optimization (\method w/o FDO), which includes hybrid rasterization and type exchange but no frequency regularization. \method w/ DWT applies frequency losses directly as in Eq.(\ref{e:loss}), yielding only marginal gains due to unresolved gradient conflicts (Figure\ref{fig:conf}). A simple alternative is to mask out conflicting frequency gradients—for example, ignoring high-frequency gradients for 2D Gaussians. We denote this variant as \method w/ DWT + mask. While masking helps reduce conflicts, it may discard useful gradient signals. In contrast, our full method achieves the best performance by leveraging gradient projection to suppress only the conflicting components while retaining informative gradients. For theoretical background on gradient projection and conflict resolution, we refer readers to~\cite{yu2020gradient}.

\begin{table*}[h]
    \footnotesize
    \centering
    \setlength{\tabcolsep}{8pt}
    \caption{Comparison of inference efficiency. We report the average FPS in each dataset.}
    \label{tab:ood}
    \resizebox{0.5\linewidth}{!}{
    \begin{tabular}{l | ccc}
    \toprule
    Method & LLFF & Tanks\&Temples & Mip-NeRF360.  \\
    \midrule
    
    3DGS & 323 & 158 & 145 \\
    2DGS & 187 & 59 & 76 \\
    GaussianPro & 308 & 166 & 121 \\
    \method & 268 & 125 & 104 \\
    \bottomrule
    \end{tabular}
    }
    \label{tab:infer}
\end{table*}

\section{Inference Efficiency}
\label{sup:infer}
As shown in Table~\ref{tab:infer}, we compare the rendering efficiency of different methods in terms of frames per second (FPS). During training, the number of parameters significantly impacts performance, as backpropagation and parameter updates are computationally expensive. In contrast, inference efficiency is primarily determined by the rasterization strategy. 3DGS-based methods, including GaussianPro, employ affine projection-based rasterization, which is efficient but less accurate, resulting in higher FPS at inference. Since both 3DGS and GaussianPro use the same projection-based rasterization pipeline, the difference in their inference speed mainly arises from model size—that is, the number of Gaussians used. While the number of primitives affects performance, its influence remains moderate given the similar scale of models.

In contrast, 2DGS adopts a ray–splat–intersection rasterization pipeline, which provides improved geometric accuracy but is more computationally intensive, resulting in slower rendering. \method integrates both 2DGS and 3DGS rasterization strategies in a hybrid manner, achieving a favorable balance between accuracy and efficiency. While \method’s FPS is slightly lower than that of 3DGS, it remains significantly faster than 2DGS. Additionally, \method benefits from a shorter training time than 3DGS, owing to its reduced model size and more effective optimization dynamics.

\section{Discussion}
\label{app:diss}
\textbf{Broader Impact.}  
This work introduces an exchangeable hybrid Gaussian splatting framework that improves the trade-off between geometry accuracy and appearance fidelity in neural rendering. By enabling flexible type adaptation and frequency-aware supervision, our method can enhance 3D reconstruction quality in both synthetic and real-world scenarios. Potential applications include autonomous driving, augmented reality, and robotics, where accurate scene geometry and photorealism are both essential. While our approach primarily targets academic benchmarks, it may inform future developments in real-time perception systems.

\textbf{Limitations.}  
The current initialization of Gaussian types is random and does not incorporate semantic or structural cues from the sparse point cloud, which may limit early-stage optimization. Incorporating semantic priors could improve convergence and final quality. Additionally, as a general-purpose representation, our method has not been explicitly tested under extreme conditions such as low-light environments, highly reflective surfaces, or scenes with significant transient content. Evaluating and adapting the framework to such challenging scenarios may further demonstrate the robustness and versatility of the hybrid representation.
